\documentclass[preprint,12pt]{elsarticle}

\usepackage{amssymb}
\usepackage{amsmath}

\journal{Measurement}
\DeclareMathOperator*{\argmax}{arg\,max}

\usepackage{graphicx}
\usepackage{multirow}
\usepackage{bigdelim}
\usepackage{subfigure}
\usepackage{indentfirst}
\usepackage{booktabs}
\usepackage{url}
\usepackage{rotating}
\usepackage{caption}
\usepackage{subcaption}
\usepackage{cleveref}
\usepackage{textcomp}
\usepackage{caption}
\usepackage{svg}
\usepackage{pgf-pie}
\usepackage{tikz}
\usetikzlibrary{shapes}
\usepackage{array}
\usepackage{arydshln}
\usepackage{xcolor}

\usepackage{placeins}

\begin{document}

\begin{frontmatter}



\title{AutoML for Multi-Class Anomaly Compensation of Sensor Drift}





\author{Melanie Schaller} 
\author{Mathis Kruse}
\author{Antonio Ortega}
\author{Marius Lindauer}
\author{Bodo Rosenhahn}

\affiliation[tnt]{organization={Institute for Information Processing (tnt), Leibniz University Hannover},
            country={Germany}}

\affiliation[usc]{organization={Department of Electrical and Computer Engineering, University of Southern California},
            country={U.S.}}

\affiliation[luhai]{organization={Institute for Artificial Intelligence, Leibniz University Hannover},
            country={Germany}}

\begin{abstract}
Addressing sensor drift is essential in industrial measurement systems, where precise data output is necessary for maintaining accuracy and reliability in monitoring processes, as it progressively degrades the performance of machine learning models over time. Our findings indicate that the standard cross-validation method used in existing model training overestimates performance by inadequately accounting for drift. This is primarily because typical cross-validation techniques allow data instances to appear in both training and testing sets, thereby distorting the accuracy of the predictive evaluation. As a result, these models are unable to precisely predict future drift effects, compromising their ability to generalize and adapt to evolving data conditions.
This paper presents two solutions: (1) a novel sensor drift compensation learning paradigm for validating models, and (2) automated machine learning (AutoML) techniques to enhance classification performance and compensate sensor drift. By employing strategies such as data balancing, meta-learning, automated ensemble learning, hyperparameter optimization, feature selection, and boosting, our AutoML-DC (Drift Compensation) model significantly improves classification performance against sensor drift. AutoML-DC further adapts effectively to varying drift severities.
\end{abstract}

\begin{graphicalabstract}
\begin{center}
    \centering\includegraphics[width=.99\textwidth, trim=0cm 2cm 0cm 2cm, clip]{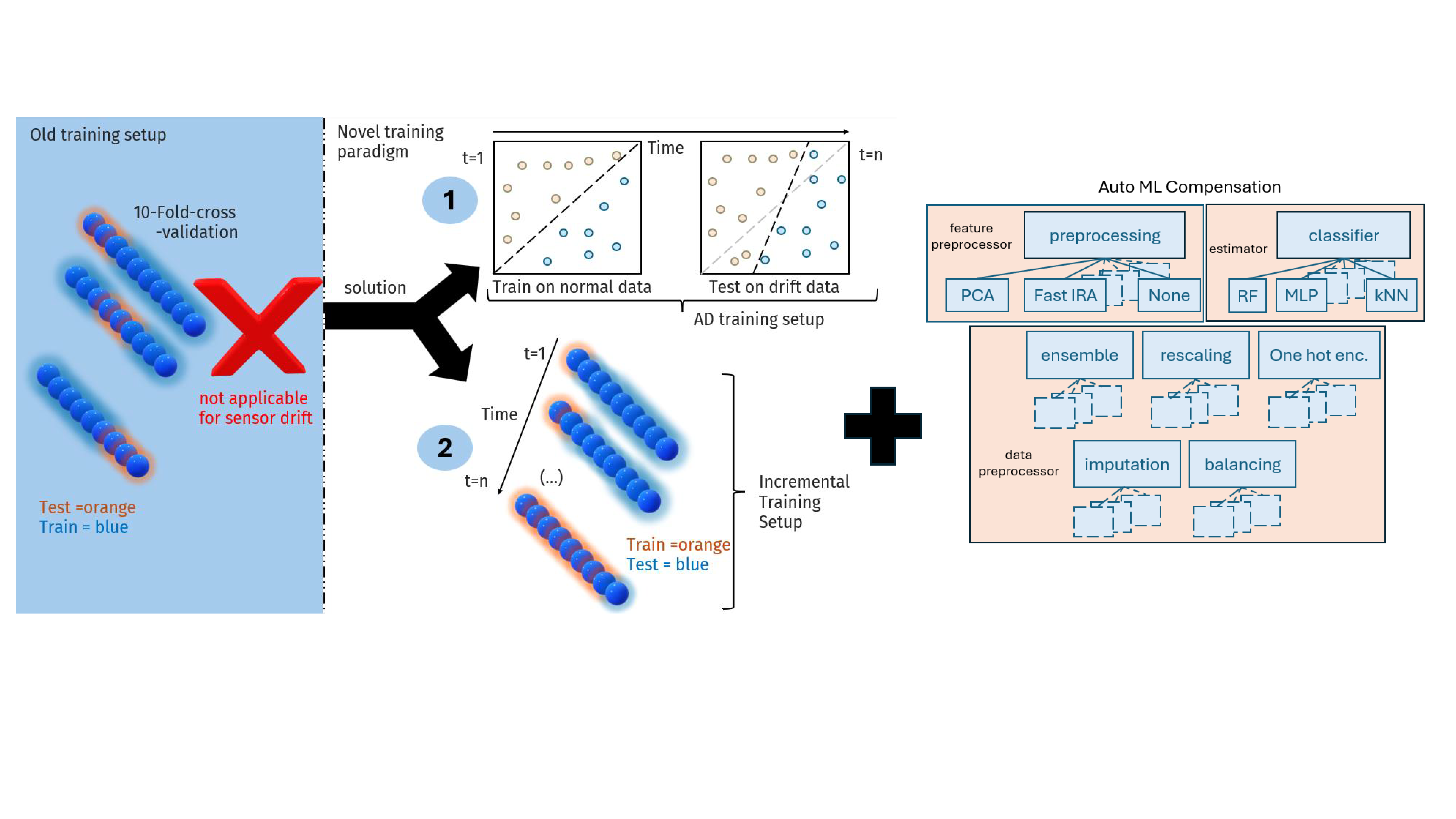}
    \captionof{figure}{Traditional training setups are inadequate for learning and compensating sensor drift (left). Our novel training paradigm, encompassing two new setups (marked with 1+2, middle), enables the network to learn drift dynamics during training. By integrating adapted AutoML techniques (right), including feature and model selection as well as hyperparameter optimization, early-stopping, and meta-learning, we prevent overfitting. This approach achieves a new state-of-the-art AutoML-DC model for sensor drift compensation.}
    \label{fig:graph_abstract}
\end{center}%
\end{graphicalabstract}

\begin{highlights}
    \item Our analysis demonstrates that the conventional training configurations are suboptimal in learning and compensating for sensor drift. Thus, we propose a novel sensor drift compensation learning training paradigm that closely matches real-world scenarios.
    \item Our findings further indicate, that AutoML techniques along with the proposed training paradigm enable effective drift adaptation to evolving levels of drift severity and complex drift dynamics in sensor measurements.
    \item By utilizing meta-learning, AutoML-DC starts from initial configurations based on prior data, lowering the extensive data requirements normally needed for training neural networks or ensemble models.
    \item We make use of AutoML techniques to enhance model robustness (see standard deviation) by combining multiple models to capture diverse data patterns, optimizing feature selection, and preventing overfitting through smart training termination.
    \item We conduct extensive benchmarking experiments against existing models and highlight the significant accuracy improvements realized when adopting AutoML-DC in practical drift compensation scenarios in industrial measurements.
\end{highlights}


\end{frontmatter}

\begin{center}
    \centering
    \captionsetup{type=figure}
    \includegraphics[width=.99\textwidth]{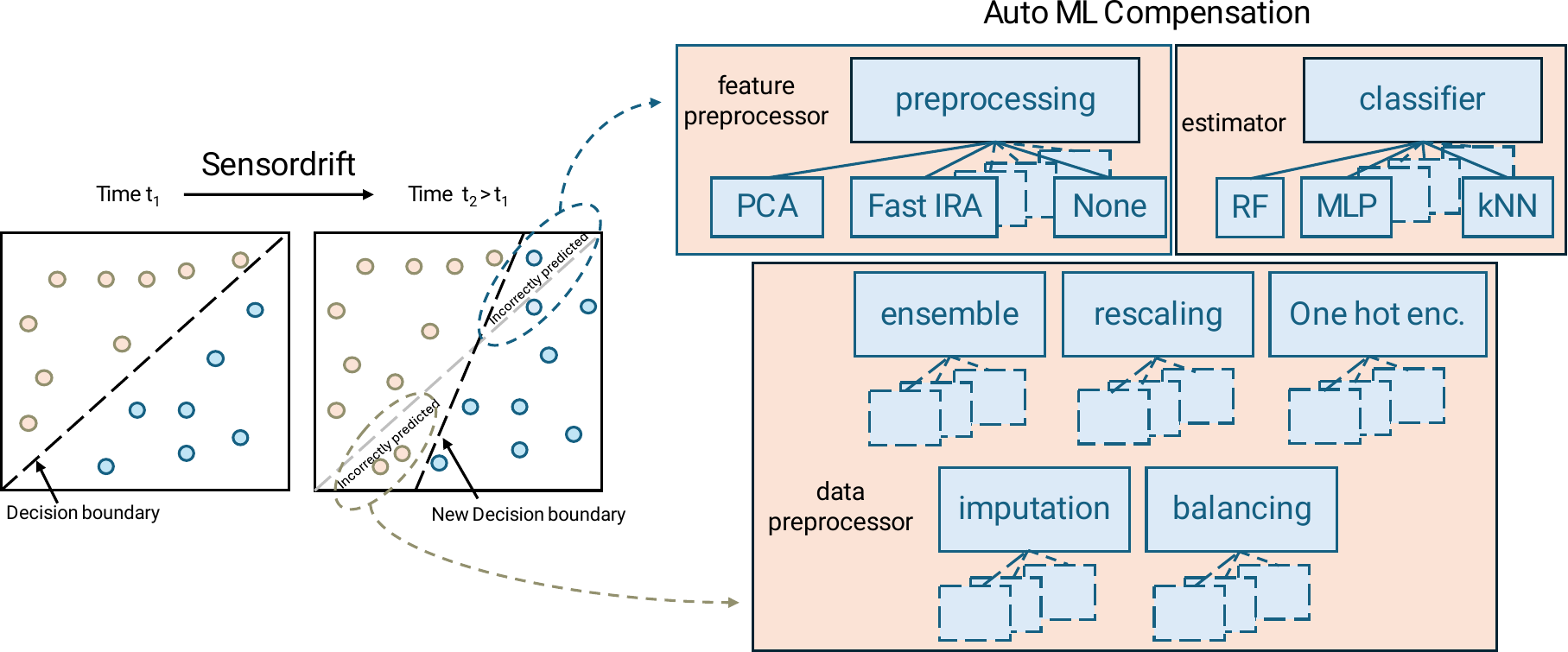}
    \captionof{figure}{Visualization of the decision boundary shift due to sensor drift and the associated incorrect prediction (left) and the usage of AutoML techniques for drift compensation (right).}
    \label{fig:teaser}
\end{center}%

\section{Introduction}\label{sec.intro}
Sensor drift is prevalent in industry~\cite{margarit2022cross}, autonomous driving~\cite{shen2020drift}, and intelligent systems with integrated sensors~\cite{wadinger2024adaptable}. In these use cases, where decision making is based on the real-time accuracy of measurement systems, sensor drift poses significant practical challenges.
This phenomenon occurs due to factors such as poisoning or environmental changes~\cite{bourgeois2003use}, sensor aging~\cite{holmberg1996drift}, and mechanical wear~\cite{yuyan2015fault}, leading to progressively inaccurate sensor readings. These inaccuracies impact machine learning (ML) models by introducing variability in input data, which compromises the accuracy and reliability of the model~\cite{godwin2009accuracy}. 
While, in general, drift is characterized by observations inconsistent with data used for training, we note that there are scenarios where the patterns of temporal change that produce sensor drift are predictable. For example, sensor aging may lead to decreased sensitivity, that is, the same ambient conditions lead to smaller sensor readings.
This paper uses automated machine learning (AutoML) techniques to develop a new training paradigm and a sensor drift compensation~\cite{WANG2025115573} solution for this type of scenario, where the patterns of identifiable temporal drift are predictable to some extent. 
Specifically, we assume that the relationship between time and changing sensor behavior can be learned in major parts as a function. This function is intended to capture the drift dynamics with linear and non-linear parts (see \Cref{sec:formalisation} and \Cref{sec:drift_linearity}) and extrapolate to future unseen data.
This allows our method to learn from the drift observed in the training data and enhance the capability of models to adapt and maintain accuracy despite sensor drift, and it could also be used for drift self-calibration in sensor measurements~\cite{AHMAD2024115158}.
Although conventional techniques use random subsets of data for training/validation, we propose two training strategies to evaluate different aspects of sensor drift compensation. 
The first training strategy involves a novel sensor drift compensation framework and handles drift in an anomaly detection setting. The second strategy utilizes an incremental batch learning approach to validate the integration performance of new drift patterns. 

Our research focuses on the following two key aspects:

\textbf{(1) Novel sensor drift compensation learning paradigm:}
As the first part of the novel learning paradigm, we introduce a sensor drift compensation learning strategy. 
For this strategy, our work closely relates to anomaly, or out-of-distribution, detection, where models are typically trained on normal data and evaluated on faulty data~\cite{chandola2009anomaly}. We propose a novel training setting for sensor drift compensation that can replace widely utilized 10-fold cross-validation or random sampling strategies for model training and evaluation~\cite{VERGARA2012320}.  
Instead, we assume an increasingly severe temporal drift will occur in the training data.
Thus, the training data represents initial drift states, and the trained model is expected to learn to compensate for observed (more severe) drift present at later stages in the test data, which can be seen as a variant of the drift adaptation task~\cite{Definition_Drift_Adaptation}. 
Most importantly, we assume that the sensor drift is separable from the data, making it possible to compensate for and reconstruct the original data from noisy measurements. However, we will show that the complex drift dynamics will make explicit drift modeling difficult, leading us to propose our model, which implicitly compensates for the sensor drift effects.

Motivated by Suárez-Cetrulo et al.~\cite{SUAREZCETRULO2023118934}, we introduce an incremental batch learning strategy as the second part of our novel training paradigm. This approach is particularly well-suited for environments where sensor data are received in a streaming fashion, and a continuous adaptation to new information is of interest. In this second learning setup, the model continuously ingests batch data, adjusting its parameters to account for drift and other variables in the environment. This ongoing learning process improves the model's ability to adapt to new patterns and anomalies, resulting in a more robust model performance even as the underlying data distribution evolves. 

\begin{figure}
    \centering
    \includegraphics[width=0.99\linewidth]{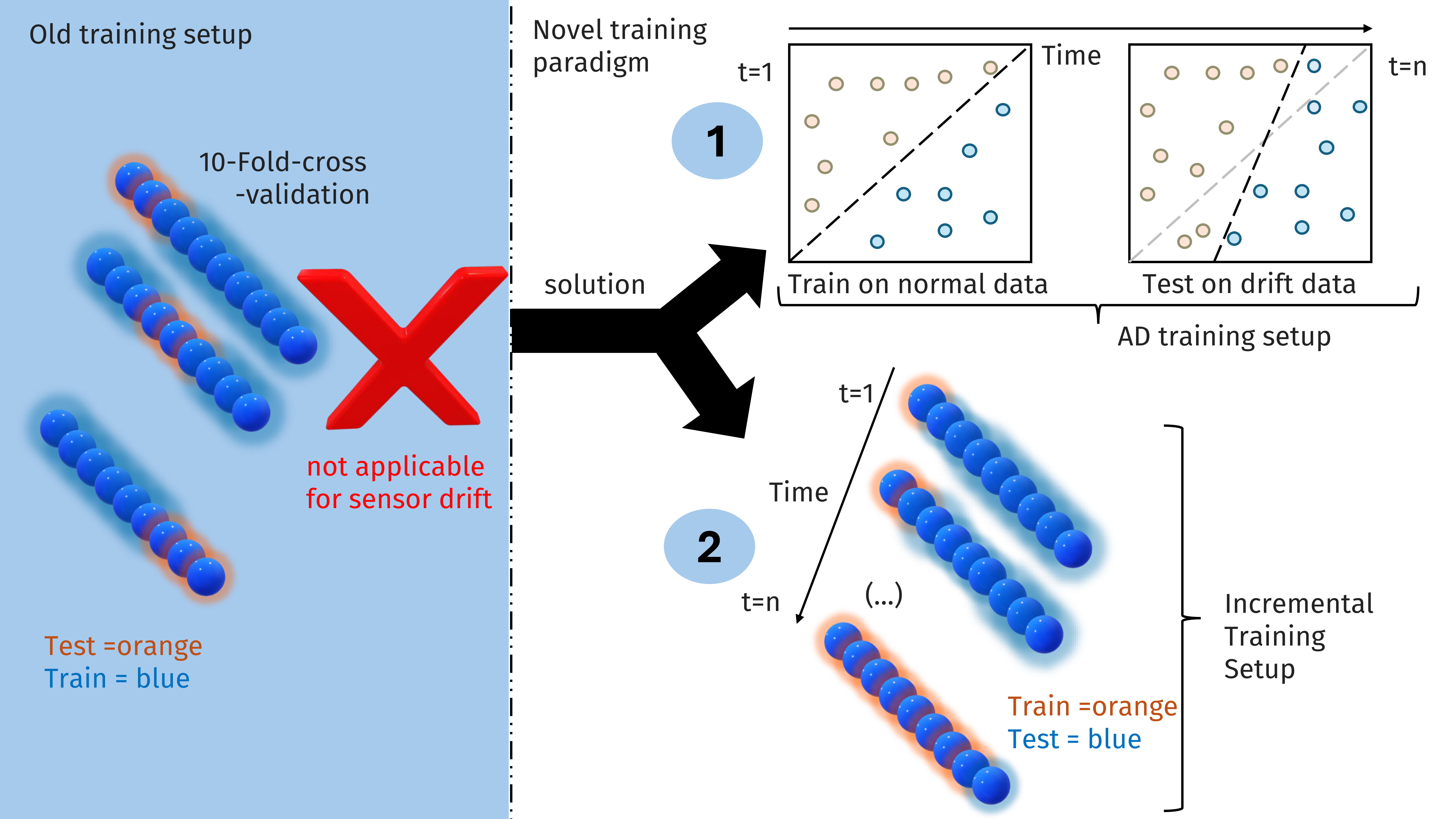}
    \caption{Vizualisation of the two-fold training paradigm on the right side vs. the traditional training paradigm on the left side.}
    \label{fig:training_paradigm}
\end{figure}

Our framework assumes an iterative learning process. The model updates its understanding of the data distribution in batches, ensuring that newly observed drift patterns are promptly integrated into the predictive framework. This method minimizes the need for large-scale retraining episodes, making it suitable for contexts that demand low-latency responses.
The combination of anomaly detection and batch learning techniques (see \Cref{fig:training_paradigm}) in one novel learning paradigm ensures the model remains vigilant to out-of-distribution events while continually refining its predictions based on the batch data.

Within different experiments, we demonstrate that several existing methods for sensor drift compensation are ineffective in learning the drift and, therefore, fail when the conventional validation setting is slightly modified.  
We demonstrate that previously published approaches cannot adequately compensate for the drift effect because of their unrealistic training setups. Existing methods typically learn average models that assume a uniform distribution across all data, making them less effective in the presence of drift. In contrast, our approach learns a static model from the initial batches and explicitly captures the drift dynamics by analyzing the differences between consecutive batches. This enables us to extrapolate the drift behavior to unseen data, providing a more accurate solution to sensor drift.

\textbf{(2) AutoML Drift Compensation:}
We show that AutoML techniques can improve the results on the drift compensation task in our proposed setting by about 16\% compared to all other benchmarking models. In our context, drift compensation means that inherent sensor drift is present in the data, which can be compensated for using AutoML techniques. 
This ensures that the model's predictions remain accurate to a higher degree despite changes in the data distributions. This is achieved by combining various models and different preprocessing and hyperparameter settings that can learn different aspects of the drift.
We showcase experimentally that employing AutoML techniques, such as automated ensemble learning with varying model weights, automated feature preprocessing, optimization of hyperparameters, boosting and imputation strategies, as well as others visualized in~\Cref{fig:teaser}, allows for the learning of anomaly patterns within the training data. This enables the model to extrapolate from smaller drift effects to increasingly pronounced anomalous effects, thus compensating for sensor drift. 

We summarize our contributions as follows:
\begin{enumerate}
    \item We demonstrate that the commonly used training settings are flawed in learning and compensating for sensor drift
    \item We propose a novel sensor drift compensation learning training paradigm that closely matches real-world scenarios.
    \item Our findings indicate that AutoML techniques\footnote{\url{https://anonymous.4open.science/r/AutoML_Drift_Compensation-466F/Readme.md}}, along with the proposed training setting, enable effective drift adaptation to evolving levels of drift severity and complex drift dynamics. 
\end{enumerate}

\section{Related Work}\label{sec:related}
\paragraph{Automated Machine Learning (AutoML)}
Machine Learning has succeeded in countless applications~\cite{TangWangWuChenPangSunFengWang+2023, TangWangWuLiuZhangZhuChenSunWang+2022+849+872,Schaller2023LiquorHGNNAH, schaller2024modeconvnovelconvolutiondistinguishing}, raising the demand for automated and streamlined solutions. The field of AutoML, which aims to find well-performing models automatically, has been receiving increased attention~\cite{AutoML:SMAC}. To ensure the flexibility and robustness needed for our sensor drift task, different AutoML techniques are applied, such as dynamic feature selection, model tuning, and adaptation to new drift patterns without extensive manual intervention, addressing the limitations of prior approaches that require fixed models and extensive domain knowledge.

Several frameworks, such as SMAC3 or auto-sklearn implement these techniques and are used in many different use cases~\cite{AutoML:auto-sklearn, AutoML:SMAC, AutoML:CASH}. Many AutoML frameworks resort to approaches such as Bayesian optimization to guide the non-trivial search for strong hyperparameters given a specific model~\cite{AutoML:SMAC, AutoML:auto-sklearn}. The problem of algorithm selection (AS) aims to find the most suitable algorithm for a given task. Other fields, such as Neural Architecture Search (NAS) aim to find new neural network architecture and topologies, to solve new tasks~\cite{AutoML:NAS}.

\paragraph{Drift Compensation}\label{sec:drift_compensation}
Prior drift compensation methods can be categorized into five types: component correction, adaptive methods, sensor signal preprocessing, attuning methods, and machine learning approaches.

\emph{Component correction methods} use methods such as Principal Component Analysis (PCA) or Independent Component Analysis (ICA) to identify and eliminate drift components~\cite{artursson2000drift, DINATALE2002158}. For dynamically evolving data sets, which regularly change due to drift, these component correction methods would need continual retraining to consider current statistics—making them labor-intensive and inefficient in comparison to systems designed to compensate dynamically without regular re-training. Furthermore methods like PCA, primarily a linear dimensionality reduction method, assumes that the main variability in the data can be captured in a reduced orthogonal space. This works well for stable datasets but can underperform if variability is erratic, time-dependent, or non-linear. ICA finds components that are statistically independent, which might not align with how drift manifests over time. Drift often appears as correlated sequential data changes not fully captured by static independence assumptions. In comparison our AutoML Drift Compensation framework allows a flexible adaptation by learning patterns, updating as the data evolves without the need for constant retraining from scratch, unlike static PCA/ICA frameworks.


\emph{Adaptive methods} include evolutionary algorithms that optimize a multiplicative correction factor for incoming samples. These algorithms, like the one proposed by Di Carlo et al.~\cite{DICARLO20111594}, continuously adapt the correction factor through linear transformations within a restricted time window. Although Evolutionary algorithms can find optimal solutions within complex, high-dimensional spaces, the multiplicative correction factor assumes drift can be corrected through simple linear scaling, which might not suffice for nonlinear drift patterns. The focus on short-term optimization can also lead to overfitting to noise or transient anomalies in the data hindering adaptation to sustained nonlinear drift dynamics. In comparison our AutoML ensemble methods might capture multi-faceted drift patterns by combining models that individually address different components of the drift. AutoML-DC can also include model evaluation strategies that balance fitting the data while avoiding over-adjustment to noise.

\emph{Preprocessing methods} involve baseline manipulation and filtering strategies. Baseline manipulation transforms sensor signals based on initial values using differential, relative, or fractional transformations. Filtering strategies, such as the Discrete Wavelet Transform (DWT), mitigate drift by discarding low-frequency components associated with drift and reconstructing the signal from the remaining components~\cite{HUI2003354}. Nevertheless, preprocessing techniques generally assume that drift patterns, such as baselines or low-frequency components, remain constant over time. This constancy allows them to calibrate and correct the data based on fixed parameters. Thus it is not useful for dynamically adjusting to new drift patterns or evolving drift like in our usecase, as it is used in a rather static manner. With the Auto-ML DC model, we instead combine and learn different preprocessing parameters dynamically according to the temporal drift patterns, that are learnable. Choosing and combining preprocessing strategies alongside model configurations also allows more immediate responses to evolving drifts. 

\emph{Attuning methods} aim to correct drift components without relying on calibration samples, instead deducing drift directly from training data. Orthogonal Signal Correction (OSC) is one such method, which removes non-correlated variance in sensor-array data~\cite{PADILLA201028}.Methods like Orthogonal Signal Correction (OSC) remove components orthogonal to the drift, thus eliminating the non-correlated variance in the data set. Thus, They rely on previously seen drift effects being representative for current and future drift compensation. Attuning methods often rely on the assumption that drift manifests in identifiable components (e.g., orthogonality) that are separated and compensated. Unlike attuning methods that are preset to correct only previously identified drift components, AutoML-DC can learn from broader, potentially evolving drift patterns within and beyond initial training data, which is shown in the extensive experiments with different training strategies. In cases where drift doesn’t appear as (e.g. orthogonal) component, AutoML-DC might also recognize shifts in sensor behavior dynamically across the operational data range.

\emph{Machine learning approaches} initially focused on adaptive drift correction using neural networks~\cite{kohonen}. These methods, however, demand a substantial number of training samples and are tightly integrated with specific algorithms. To address flexibility, Vergara et al.~\cite{VERGARA2012320} introduced an ensemble drift compensation method, utilizing features such as steady-state and normalized responses and employing classifiers like SVMs. 
Various machine learning models have been proposed, often using random train-test splits or cross-validation, and they are thus trained in a setting other than our proposed drift compensation setting. While Machine Learning models typically require large volumes of training data to accurately model the drift, especially when using neural networks or complex models to ensure convergence and generalization, we use meta-learning strategies to start from an informed initial configuration, reducing the need for exhaustive training data. While methods like those in Vergara et al.'s ensemble often rely on pre-selected models and handcrafted feature sets, AutoML-DC offers dynamic adaptability by exploring a broad range of models and hyperparameter settings automatically, selecting combinations that best capture the current drift patterns.

\section{Formalisation of the Drift Compensation Problem}\label{sec:formalisation}

In real-world applications, sensors often operate over extended periods, leading to aging and degradation. This degradation is commonly referred to as sensor drift, induced by elusive dynamic processes such as poisoning, aging, or environmental variations \cite{Falasconi12, HOLMBERG1997185} and has to be compensated by machine learning models, that are employed to monitor sensory systems.
The drift compensation problem can be formulated as follows. 
Let $T_1, T_2, \ldots, T_K$ denote time-series data across \(K\) batches, organized chronologically.
Each time series \(T_i\) is defined as \(T_i = \{x_{ij}\}_{j=1}^{N_i}\), where \(x_{ij}\) represents the feature vector of the \(j\)-th sample in Batch \(i\), and \(N_i\) is the number of samples in Batch \(i\).
The sensor drift issue arises when the feature distributions of \(T_2, \ldots, T_K\) deviate from that of \(T_1\).
Consequently, a classifier trained on labeled data from \(T_1\) exhibits degraded performance when tested on \(T_2, \ldots, T_K\) due to diminished generalization caused by drift, which needs to be compensated.
The mismatch in distribution between \(T_1\) and \(T_i\) becomes irregularly more pronounced with increasing batch index \(i\) (\(i > 1\)) and aging.

\section{Formalisation of the Anomaly Compensation Task within AutoML}\label{sec:formalize_anomaly}
The goal is to train a classifier \(f\) using the labeled data from the first $k_{train} < K$ batches, with $\mathcal{D}_{train}=\{1, \ldots, k_{train}\}$, in a supervised manner.
The classifier is trained to predict class labels \(C_{ij}\) based on the feature vectors \(x_{ij}\).
The classifier has to learn both the normal data as well as initial drift patterns from the first few batches and generalize them to later batches where increased drift severity is observed.
We argue that generalizing from the initial drift effects to the more pronounced drifts in later batches is a more realistic and more challenging setting.
%
The trained classifier \(f\) is then tested on the last $k_{test} = K - k_{train}$ batches, i.e. $\mathcal{D}_{test}=\{k_{train} + 1, \ldots, K\}$.

To gain enough flexibility to compensate for all drift effects, we model our classifier $f$ as an ensemble of known models, such as MLPs or Random Forests, and optimize it as an \emph{algorithm selection and hyperparameter optimization problem (CASH)}~\cite{AutoML:CASH}. We determine the set of algorithms for the ensemble out of a pool of algorithms $\mathbf{A},$ with each $a_i \in \mathbf{A}$ having its own hyperparameter space $\mathbf{\Lambda_i} \in \mathbf{\Lambda}$.
Searching for the best-performing model becomes the optimization problem
\begin{equation}
    (a^*, \lambda^*) \in \argmax_{a_i \in \mathbf{a}, \lambda \in \mathbf{\Lambda_i}} c(a_i, \lambda),
\end{equation}
where $a^*$ denotes the optimal choice of model and $\lambda^*$ the respective choice of hyperparameters. The cost function $c(a_i, \lambda)$ quantifies the performance of the current model $a_i$ with some hyperparameter choice $\lambda$. In our case, $c$ is modeled using the F1-score, while we also track metrics such as precision and recall. Using the $k$ best-performing models determined by the optimization problem above, an ensemble is built to make predictions more robust against sensor drift.

In our paper, we optimize this problem using the auto-sklearn framework~\cite{AutoML:auto-sklearn}, which also optimizes the choice of feature pre-processing, such as different embeddings, PCA or other encodings.
To navigate the search space more efficiently, trading off exploration and exploitation, Bayesian optimization methods are used to guide the search.
Utilizing results from meta-learning, the models are instantiated using initial instantiations pre-computed by auto-sklearn, which are determined using carefully selected and empirically found meta-features.
The final ensemble is built with ensemble selection techniques and validated on a hold-out set~\cite{ AutoML:ensemble_selection, AutoML:auto-sklearn}.

\section{Dataset Description}\label{sec:dataset}
To the best of our knowledge, the dataset by Vergara et al.~\cite{VERGARA2012320} is the only dataset that fully represents the sensor drift problem in a practical setting. This dataset is particularly valuable for our research because it captures the complexities of sensor drift in a real-world industrial environment, where such issues frequently occur. Other sensor drift datasets, such as those from IntelLab~\cite{de2016benchmark}, Santander~\cite{de2016benchmark}, and SensorScope~\cite{de2016benchmark}, primarily involve synthetic drift, which does not fully capture the nuanced challenges presented by natural sensor drift. Specifically, Vergara et al. curated a dataset featuring responses from a sixteen-element array of metal-oxide semiconductor gas sensors in a 60 ml test chamber. Various odorants, including ammonia, acetaldehyde, acetone, ethylene, ethanol, and toluene, that represent the multi-classes, were injected into the chamber and measured at a constant flow rate of 200 ml/min. The sensors operated at 400 °C, heated by an external DC voltage source. Resistance time series with a 100 Hz sampling rate, were collected over 36 months, with a deliberate 5-month gap to induce contamination. The dataset contains a total of 13,910 recordings and is introduced as sensor drift dataset with increasing drift severity over time.

\subsection{Batch distributions and dataset structure}\label{sec:batch_distribution}

According to the setting's definition above, we divide the used data set into $K=10$ batches, subdivided into $k_{train} = 5$ training and $k_{test}=5$ test batches. All runs in the code have been repeated ten times to see the robustness and significance of the results and the standard deviation has been calculated. 
In order to conduct a fair comparison of all models the hyperparameters of all models have been optimized due to their specific conditions. As all implemented benchmarking models have different specifications, we track the full set of tuned hyperparameters in the appendix due to capacity reasons. 

Since most samples are recorded in later batches, up to the 16th month or Batch 5, we extended the data inclusion up to the fifth batch for training. Thus, the training dataset contains 3633 samples out of 10277. This decision was driven by the already substantial imbalance in the dataset~\cite{Dennler2021DriftIA}. Further data set descriptions are found in~\ref{sec:appendix_a1} and ~\ref{sec:appendix_a2}.

\section{Novel Sensor Drift Training Paradigm}\label{sec:training_paradigm}
As part of our novel learning paradigm, we address the sensor drift compensation challenge using a two-fold strategy that combines anomaly detection principles with an incremental batch learning approach. This paradigm is designed to validate the model's robustness and adaptability in the face of dynamic sensor drift.

First, we utilize a learning approach inspired by anomaly detection, training models on initial, drift-free data. This data serves as a baseline for adapting to intensified drift conditions, allowing the model to implicitly manage the complex, non-linear dynamics of sensor drift without explicit modeling.

Second, we implement an incremental batch learning strategy for real-time data environments. This approach enables the model to continuously adjust parameters in response to new data, ensuring robust performance against evolving drift characteristics. This method minimizes the need for comprehensive retraining, allowing the model to integrate new drift patterns and maintain accuracy. The incremental addition of batches simulated a real-world scenario where a model is periodically retrained with new data and was motivated from~\cite{Lu_2018}.

\section{Evaluation Metrics}\label{evaluation_metrics}

To evaluate the performance of our models in compensating for sensor drift, we employ several key metrics: Precision, Recall, F1-Score, Accuracy, and the Area Under the Receiver Operating Characteristic Curve (AUC-ROC). Precision is defined as the ratio of true positive predictions to the sum of true positive and false positive predictions. It measures the accuracy of the positive predictions made by the model~\cite{powers2020evaluationprecisionrecallfmeasure}. Recall, also known as sensitivity, measures the ratio of true positive predictions to the sum of true positives and false negatives~\cite{powers2020evaluationprecisionrecallfmeasure}. The F1-Score is the harmonic mean of Precision and Recall, especially useful for imbalanced datasets~\cite{powers2020evaluationprecisionrecallfmeasure}.Accuracy is defined as the ratio of correctly predicted instances to the total instances in the dataset~\cite{hastie}.The AUC-ROC metric evaluates the model's ability to distinguish between classes across various thresholds, offering a comprehensive measure of classification performance~\cite{Fawcett}.

The choice of these metrics is motivated by their relevance to the task of sensor drift compensation. Precision and recall provide insights into the correctness and completeness of positive predictions, respectively. The F1-score balances these two metrics, accounting for class imbalances. Accuracy is used for its general assessment capability and is supplemented by AUC-ROC to ensure robust evaluation across various thresholds. These metrics should collectively enable a multifaceted performance evaluation. In our study, machine learning models are specifically designed to recognize and compensate for sensor drift over time, emphasizing accuracy and robustness across drift levels rather than immediate detection and response. Thus, we employ metrics that effectively assess how well the models manage changing data distributions due to drift. Metrics such as drift detection delay and adaptation time are more pertinent to system-level responses, where operational adjustments are critical. However, our approach focuses on optimizing model parameters for predictive accuracy amidst drift within controlled training paradigms. Thus, speed-focused metrics have been excluded.

\section{Hyperparameters of baseline models}\label{sec: hyperparameters}

In our study, we tuned the hyperparameters across a range of machine learning models to ensure fair benchmarking. For the XGBoost model, we optimized the $\emph{learningrate}$, $\emph{max-depth}$, and $\emph{n-estimators}$ hyperparameters. The Support Vector Machine (SVM) model required tuning of $\emph{C}$ (regularization parameter), $\emph{gamma}$ (kernel coefficient), and $\emph{kernel}$ type. For the Random Forest classifier, we adjusted the $\emph{max-depth}$, $\emph{min-samples-split}$, and $\emph{n-estimators}$. For Logistic Regression, the hyperparameters $\emph{C}$ (inverse regularization strength), $\emph{solver}$ (optimization algorithm), and $\emph{max-iter}$ (maximum iterations) were optimized. The Gradient Boosting model required tuning of the $\emph{learningrate}$, $\emph{max-depth}$, and $\emph{n-estimators}$. For Gaussian Naïve Bayes (GaussianNB), we focused on optimizing $\emph{var-smoothing}$. 

The Decision Tree classifier's hyperparameters $\emph{criterion}$ (splitting function), $\emph{max-depth}$, and $\emph{min-samples-split}$ were modified. The performance of the k-Nearest Neighbors (kNN) model was enhanced by selecting the optimal $\emph{num-clusters}$. For ARIMA, we used the $\emph{autoarima}$ function. The Autoencoder required adjustments of the $\emph{input-dim}$, $\emph{encoding-dim}$, $\emph{num-layers}$, and $\emph{learningrate}$. Hyperparameters for Auto-Sklearn included $\emph{per-run-time-limit}$, $\emph{ensemble-size}$, and $\emph{metalearning}$ components.

For GRU (Gated Recurrent Unit) and LSTM (Long Short-Term Memory) networks, we tuned $\emph{hidden-size}$, $\emph{num-layers}$, and $\emph{learningrate}$. The Convolutional Neural Network (CNN) was optimized with $\emph{filter-size}$, $\emph{kernel-size}$, $\emph{pool-size}$, and $\emph{dense-units}$. The Drift-Ensemble, using a "hard" voting strategy, was optimized with the Kruskal-Wallis, Shapiro-Wilk, and Mann-Whitney U tests. In order to tune other hyperparameters the "hard" voting strategy would have to be changed to a "soft" voting strategy, which was not the intention of the paper. Therefore, we left the model like it was supposed to be from the authors. For Anomaly-GAN, we fine-tuned $\emph{hidden-dim}$ and $\emph{learningrate}$. These hyperparameter optimizations have been used to ensure a fair comparison. The ranges of the optimized hyperparameters can be looked up in the repository.
The libraries are listed in the `requirements.txt` file of the provided repository. All experiments were conducted on CPUs, specifically Intel Xeon Processor with 128GB RAM, running Ubuntu 18.04. However, the code is compatible with GPU execution as well, which can be utilized depending on availability.

\section{Experimental Results}\label{sec6:experimental_results}
As discussed before, we use the well-studied sensor drift dataset proposed by Vergara et al.~\cite{VERGARA2012320}, which contains real-world sensor-drift data.
In this dataset, a certain baseline drift can be observed, but also additional short-term as well as long-term drift effects~\cite{Dennler2021DriftIA}. Therefore, the machine learning models employed for classifying multiple classes must be able to learn the basic patterns of a hybrid form of sensor drift (see \Cref{fig:dennler_2}) at early stages and accurately learn and predict the subsequent higher levels.
Rather than relying on random sampling or ten-fold cross-validation~\cite{VERGARA2012320}, we aim to train the model on our proposed training paradigm. In the first benchmarking results section, we use the Anomaly detection strategy for training. Subsequently, the model should demonstrate proficiency in predicting the accurate classes for the ensuing five batches, characterized by distinct distributions from the initial batches due to high drift severity. Consequently, the models are tasked with learning the distinctive sensor drift patterns present in the initial batches, enabling them to forecast the correct classes for the subsequent unseen batches.
As comparison we also showcase the results on the originally proposed training setup for selected models to proove our claim.
To further evaluate the classification performance, we also document the AUC-ROC Scores for all benchmarking models.
We also conducted a drift linearity test to distinguish between drift effects. 
Afterwards we also compare the decision boundaries of Random Forest as most frequently used model in the AutoML-DC ensemble against Support Vector Machine (SVM) boundaries with RBF-Kernel.
To further investigate the robustnes of the models we also compare the standard deviation and mean accuracy values for all benchmarking models.
Afterwards we use the second learning strategy of our learning paradigm to show the capability of the models to adapt to temporal changes of drift effects.
Lastly we conduct extensive experiments on the effects of AutoML techniques within the AutoML-DC model to showcase the single effects within an ablation study.
To ensure comparable results, the number of epochs and batch sizes were kept constant across all experiments, while all hyperparameters were specifically tuned according to the specific model conditions.

\subsection{Benchmarking results for sensor drift with the anomaly detection training setup}\label{sec:experiments_anomaly_detection}

Based on prior work, as discussed in~\Cref{sec:related}, we choose the most frequently used models that have been implemented on the dataset. Since Random Forest~\cite{ijaz2020recursive} showed good results in other studies, we explore other decision tree-based models~\cite{habib2019classification} and Gradient Boosting~\cite{pareek2021smart}, to assess their impact in comparison. 
Next, we compare these results against Kernel methods like SVM~\cite{zhao2019sensor} with RBF Kernel and a Gaussian Naive Bayes Model~\cite{saeed2020machine}.
The third group of models we choose for comparison are temporal baseline models like LSTM~\cite{zhao2019sensor}, GRU~\cite{chaudhuri2020attention} and a temporal CNN~\cite{jana2022cnn}. As a fourth group, we also investigate the performance of CatBoost~\cite{zamansky4768947swcnt} against AdaBoost~\cite{lin2019concept}, XGBoost~\cite{fan2023fault} and Bagging~\cite{sarnovsky2021adaptive}.
The fifth group of models is the instance-based learning model KNN (k nearest neighbours)~\cite{adhikari2014multiple} with an optimized number of neighbours.
As one of the review papers on this dataset \cite{9756340} stated, that spiking neural networks (SNN) could be useful to solve tasks on drift data, we included these models as the sixth group of models. 

\begin{table}[ht]
    \centering
    \caption{Benchmarking results on the proposed sensor drift compensation setting}
    \label{tab:baseline performances}
    \begin{tabular}{lcccccc}
    \toprule
    \textbf{Model} & \textbf{Precision} & \textbf{Recall} & \textbf{F1} \\
    \midrule
    Random Forest & 0.68 & 0.57 & 0.56 \\
    SVM (RBF Kernel) & 0.52 & 0.43 & 0.43 \\
    Logistic Regression & 0.57 & 0.53 & 0.50 \\
    XG Boost & 0.66 & 0.53 & 0.51 \\
    CatBoost & 0.49 & 0.54 & 0.50 \\
    KNN & 0.68 & 0.57 & 0.56 \\
    SNN & 0.16 & 0.13 & 0.11 \\
    LSTM & 0.58 & 0.61 & 0.57\\
    CNN & 0.65 & 0.62 & 0.60\\
    Decision Tree & 0.50 & 0.38 & 0.40 \\
    Gradient Boosting & 0.49 & 0.51 & 0.49 \\
    Gaussian Naive Bayes & 0.50 & 0.32 & 0.32 \\
    AdaBoost & 0.42 & 0.42 & 0.40 \\
    Bagging & 0.48 & 0.40 & 0.39 \\
    Ensemble Model & 0.29 & 0.32 & 0.29 \\
    GRU & 0.40 & 0.32 & 0.28 \\
    Anomaly-GAN & 0.21 & 0.35 & 0.26\\
    AutoML-DC (ours) & \textbf{0.77} & \textbf{0.76} & \textbf{0.76}\\
    \bottomrule
    \label{benchmarks}
    \end{tabular} \\
\end{table}
The ensemble drift compensation, that was introduced by Vergara et al. \cite{VERGARA2012320} especially on this dataset is also used as benchmarking model.
The last group is formed by a generative adversarial network (GAN) to solve the anomaly classification or anomaly detection \cite{ngo2019fence,lian2023anomaly,raturi2023novel}. For an overview of GANs used in anomaly detection tasks see \cite{noor2023generative}. Here, the discriminator score is taken to set the threshold for each class.

\Cref{tab:baseline performances} summarizes the performance metrics of the selected machine learning models. Precision, recall, and F1-score~\cite{goutte2005probabilistic} are utilized to evaluate the models, providing a comprehensive assessment of their ability to learn patterns of sensor drift. 

As the table shows, none of the benchmarking models achieved an F1 score exceeding 60\% for our proposed drift compensation setting. 
Conversely, the ensemble drift compensation of Vergara et al. \cite{VERGARA2012320} displays lower scores across all metrics, suggesting a diminished ability to accurately detect sensor drift anomalies despite the promising results of the ensemble model for the whole dataset with random sampling. The Spiking Neural Network (SNN) also exhibits relatively low precision, recall, and F1-score, indicating limited effectiveness in this context, although it was shown to work well for the random sampling strategy.
The AutoML-DC framework leverages meta-learning strategies, streamlining the hyperparameter tuning process and reducing the computational burden compared to traditional methods. This strategic approach, in combination with the other AutoML techniques, improves model configuration, leading to a consistent outperformance with an F1 score of 76\%.
We note that integrating a rigorous comparison with state-of-the-art drift compensation methods, such as adaptive and component-correction approaches, into this multi-class classification setting presents significant challenges. These methods are not specifically designed for simpler binary classification or regression tasks, which makes direct comparisons in a multi-class context complex. Specifically, each class in the dataset may experience drift at different rates and in different manners, requiring a model that can simultaneously handle complex interdependencies across multiple classes. While a binary classification approach could isolate these class-specific drift patterns, it might not capture interactions between classes. Thus, methods other than machine learning models have been excluded from the benchmarking results to guarantee a fair comparison, given that a simple adaptation of binary classification methods in terms of One-vs-Rest or One-vs-One might not be sufficient.

\subsection{Results on original training scenario as comparison}

In order to be able to compare the results of the benchmarking study on the proposed drift compensation setting with the original training strategy, we present the results for some major models trained with 10 fold cross validation on the same dataset as follows in \Cref{tab:classification_comparison}:

\begin{table}[htbp]
  \centering
  \caption{Comparison of some models of the benchmarking study on the original proposed setting (10 fold crossvalidation), which does not model the complexity of sensor drift properly in comparison to the F1 Score of our novel proposed training paradigm (=TP F1).}
  \label{tab:classification_comparison}
  \begin{tabular}{lcccc}
    \toprule
    \textbf{Modell} & \textbf{Precision} & \textbf{Recall} & \textbf{F1-Score} & \textbf{TP F1}\\
    \midrule
    Random Forest & 0.98 & 0.98 & 0.98 & \textbf{0.56}\\
    SVM & 0.99 & 0.98 & 0.98 & \textbf{0.43}\\
    Decision Tree & 0.97 & 0.97 & 0.97 & \textbf{0.40}\\
    Logistic Regression & 0.95 & 0.95 & 0.95 & \textbf{0.50}\\
    Drift-Ensemble & 0.98 & 0.98 & 0.98 & \textbf{0.29}\\
    AutoML-CD & 0.99 & 0.99 & 0.99 & \textbf{0.76}\\
    \bottomrule
  \end{tabular}
\end{table}

As the results show, traditional training methods, such as ten-fold cross-validation, are inadequate for sensor drift compensation because they often cause data leakage, especially in small datasets (see \Cref{tab:classification_comparison} F1 comparison). These methods often mix data from different time periods, which obscures the progression of drift and hinders the model's ability to learn temporal dynamics. As a result, models trained this way may perform overoptimistically well during cross-validation (as shown in \Cref{tab:classification_comparison}) but fail when exposed to data with unseen levels of drift (as shown in~\Cref{benchmarks}). 

\subsection{AUC-ROC Scores for benchmarking models}

\begin{figure*}
    \centering
    \includegraphics[height=0.75\textwidth, angle=90, trim=2cm 1cm 2cm 1cm, clip]{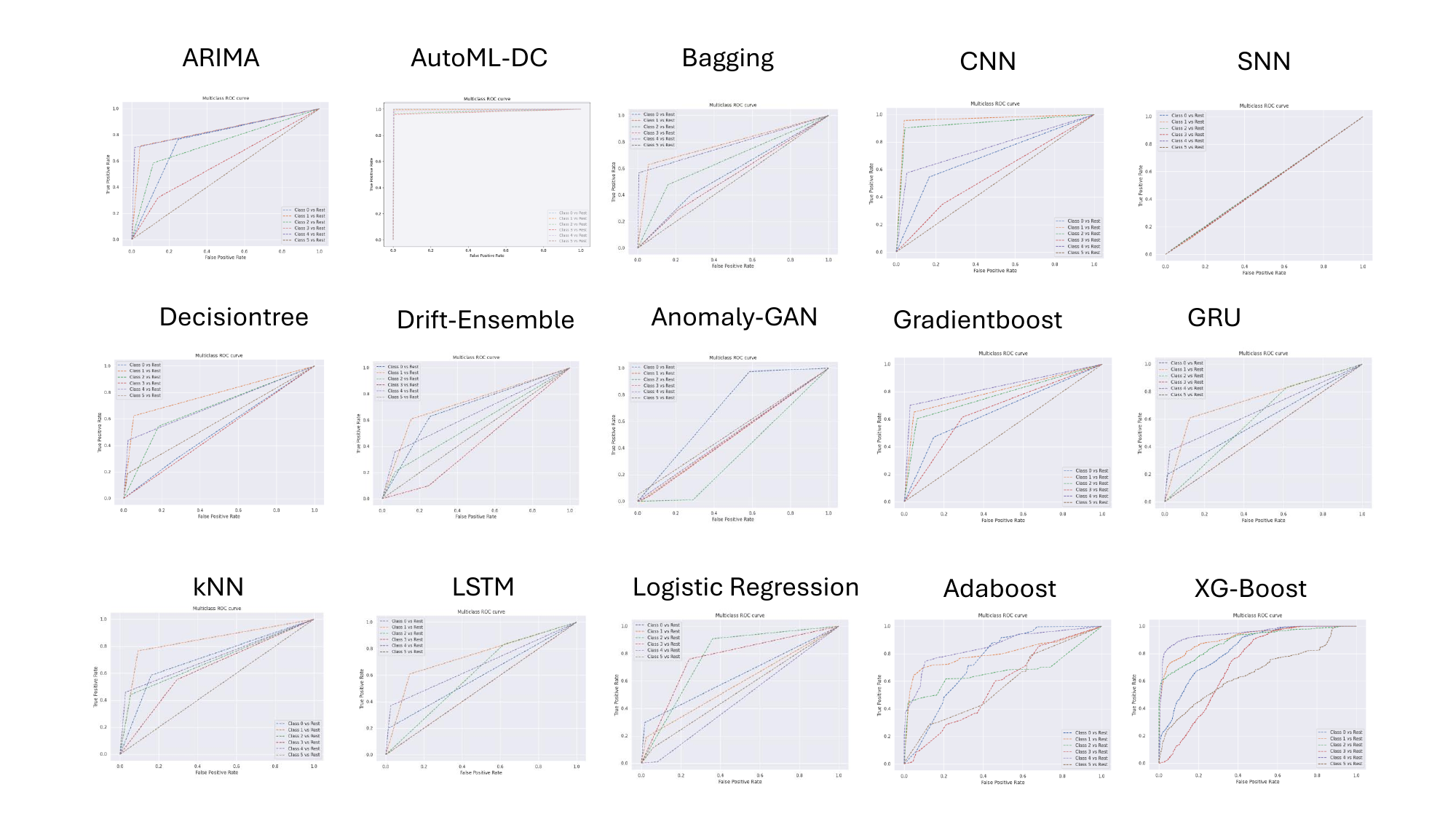}
    \caption{ROC Curves for all evaluated methods, best viewed in color and zoom in.}
    \label{fig:appendix_aurocs}
\end{figure*}

When trained in the proposed sensor drift compensation task, the AUC-ROC scores for the benchmarking models have been calculated for all classes (see \Cref{fig:appendix_aurocs}).
AutoML-DC performs exceptionally well, as indicated by its ROC curve near the top left corner. This positioning suggests that AutoML-DC achieves a high True Positive Rate with a low False Positive Rate, reflecting the best classification performance in comparison to the other models. Gradient Boost, XGBoost, Bagging, and Adaboost demonstrate strong ability to distinguish between classes, with ROC curves that approach the ideal top-left corner, indicating higher accuracy in these in comparison to the other models. Our approach prioritizes methods that align closely with the evaluation of how models adapt and maintain performance in dynamic environments. Using AUC-ROC scores, we provide a statistically robust framework to evaluate and validate the effectiveness of the model at different classification thresholds.

\subsection{Drift Linearity Test}\label{sec:drift_linearity}

We employed a support vector regression test to analyze the drift effect in the context of linearity. The primary objective of this test is to provide an initial qualitative assessment of the balance between linear and non-linear drift components using kernel function comparisons. We conducted the originally proposed ten-fold cross-validation for training and testing on the whole dataset (that is, on all batches) and compared the results of the support vector machine (SVM) with a linear kernel against the results of the SVM with an RBF kernel.

\begin{figure}[ht]
    \centering
    \includegraphics[width=0.55\textwidth]{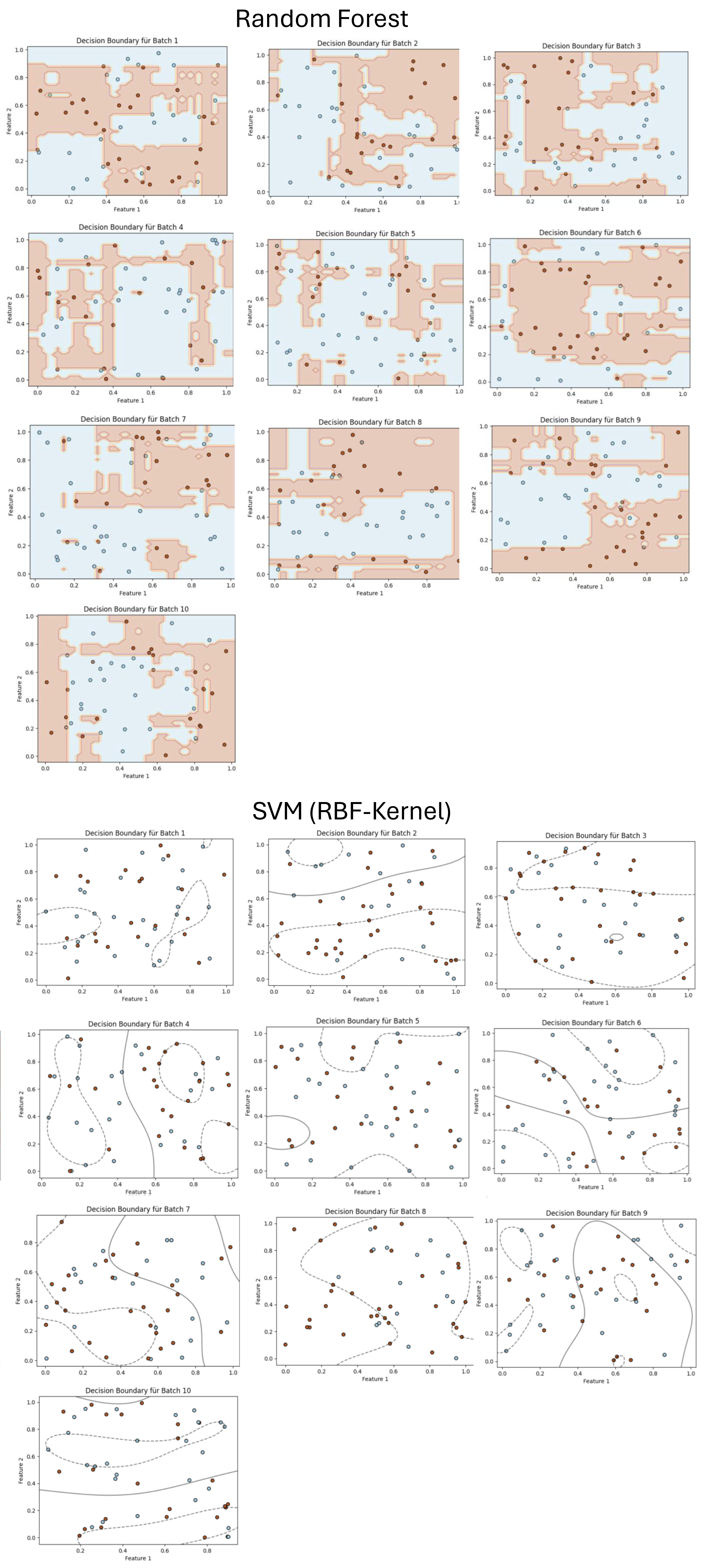}
    \caption{Decision Boundaries of Random Forest versus Support Vector Machine with RBF-Kernel, best viewed in color and zoom in.}
    \label{fig:RFSVM}
\end{figure}

The choice of kernel (linear or RBF) impacts the decision boundary of the SVM~\cite{li2018decision}. A linear kernel corresponds to a linear decision boundary in the input space. It assumes that the underlying relationship between the features and the target variable is linear. The linear kernel is effective when the data can be adequately separated by a hyperplane. The Radial Basis Function (RBF) kernel, also known as the Gaussian kernel, introduces nonlinearity by transforming the input space into a higher-dimensional space. It allows the SVM to capture more intricate relationships in the data. The RBF kernel is particularly useful when the decision boundary is complex and nonlinear, see \Cref{fig:RFSVM}.
The effectiveness of the linear kernel with an Accuracy of 0.97 implies that a significant portion of the sensor drift can be explained by linear relationships between features and classes. On the other hand, the slightly better performance of the RBF kernel with an Accuracy of 0.98 indicates that there are also additional non-linearities in the data.

\FloatBarrier 

\subsection{Decision boundaries}

As Random Forest was investigated to be a good sensor drift compensator in the drift compensation setting, we further compare the decision boundaries learned by a Random Forest to those of a Support Vector Machine with RBF-Kernel.

For this plot, 50\% of the datapoints of each batch have been taken to train the classifiers. While the Random Forest model shows quite complex decision boundaries but with almost all samples being correctly classified for the first two learned features, the SVM model with RBF kernel does not seem to learn decision boundaries, that are capable of distinguishing between the features correctly. It is even worse for the linear SVM.

\subsection{Standard Deviation and Mean Accuracy over repeated runs}

\begin{figure*}
    \centering
    \includegraphics[width=0.99\textwidth]{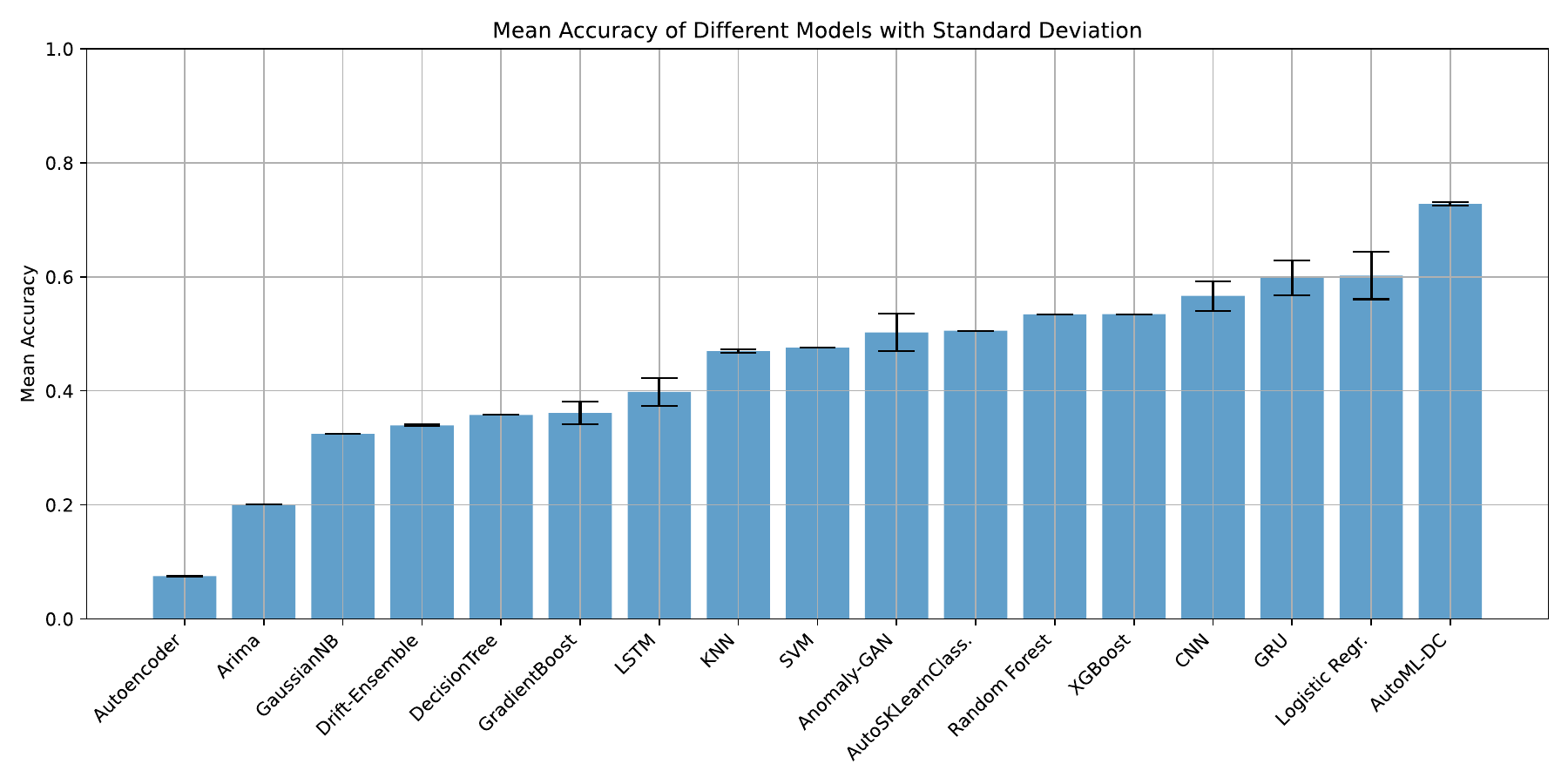}
    \caption{The mean accuracy across different models for $n=10$ runs, with their standard deviation included.\textbf{Our AutoML-DC model} consistenly outperforms competitors, while keeping a standard deviation of less than 0.3\% accuracy.}
    \label{fig:barplot_with_stds}    
\end{figure*}

The results of the reliability test are displayed in~\Cref{fig:barplot_with_stds}.


AutoML-DC achieves the highest mean accuracy with a very small standard deviation of less than 0.3\%. Through the incorporation of robust model architectures such as Random Forests with varying regularization strengths, AutoML-DC captures the diverse patterns inherent in sensor drift data. This prevents overfitting and boosts overall model reliability, as demonstrated by consistently high accuracy and low standard deviation across multiple runs. Other top-performing models include Logistic Regression, GRU (Gated Recurrent Unit), CNN (Convolutional Neural Network), and XGBoost, each showing high accuracy but with slightly higher standard deviations than AutoML-DC. In contrast, Adaboost and ARIMA have the lowest mean accuracy values, along with significant variation, which indicates lower and less stable performance. Models such as Gradient Boost, Drift-Ensemble, and Decision Tree fall in the middle range of accuracy, with moderate variation in performance.

\subsection{Results Online Learning Test}\label{sec:app_online_learning}

The following~\Cref{fig:online_learning_over_time_accuracies} illustrates model accuracy progression through the incremental batch learning strategy, emphasizing each model's capacity to seamlessly integrate new data batches over time:

\begin{figure*}
    \centering
    \includegraphics[width=0.99\textwidth]{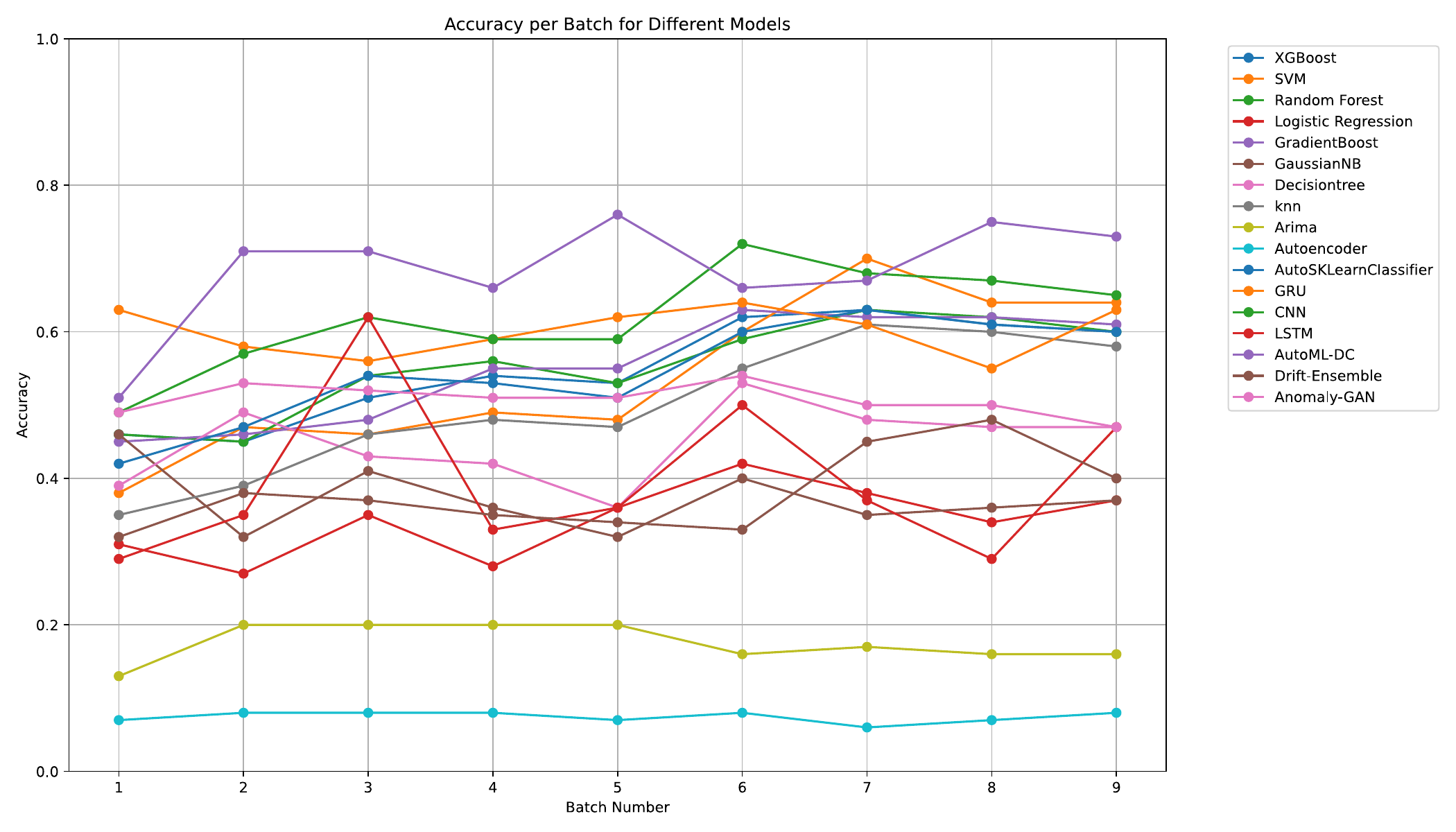}
    \caption{The performance during the Batch online Learning Test, for varying numbers of batches. Our AutoML-DC model outperforms the baseline models for almost all batches.}
    \label{fig:online_learning_over_time_accuracies}
\end{figure*}

The line chart illustrates the performance of different models during the Batch Online Learning Test across varying numbers of batches. The y-axis represents accuracy, while the x-axis represents the batch number. Each line corresponds to a specific model, with the legend on the right identifying the models by color.

Throughout most of the batches, AutoML-DC consistently outperforms the baseline models, maintaining relatively high accuracy compared to others. Other models, such as XGBoost, Random Forest, Logistic Regression, and Gradient Boost, also perform fairly well, though with more fluctuations in accuracy across batches. Some models, like Adaboost and ARIMA, show lower and more variable performance, remaining at the lower end of the accuracy spectrum. This visualization highlights AutoML-DC's ability to adapt and perform robustly in an online learning setting, as it achieves higher accuracy with minimal variability across the batches, unlike several other models that experience more inconsistency. Furthermore, we observe that for stronger drift dynamics, more of the highly non-linear Random Forests are used in the Ensemble. Additionally, the number of models in the ensemble also rises with increasing drift and difficulty.

\subsection{AutoML results on the anomaly compensation task}

\begin{table}
    \centering
    \caption{AutoML-DC Ensemble Classification for 50\% of batches}
    \begin{tabular}{cccccc}
    \toprule
    \textbf{Class} & \textbf{Precision} & \textbf{Recall} & \textbf{F1} & \textbf{Support} \\ \midrule
    1 & 0.96 & 0.94 & 0.95 & 132 \\
    2 & 0.96 & 0.99 & 0.98 & 199 \\
    3 & 0.97 & 0.94 & 0.95 & 97 \\
    4 & 0.93 & 0.93 & 0.93 & 99 \\
    5 & 0.98 & 0.98 & 0.98 & 181 \\
    6 & 0.88 & 0.79 & 0.83 & 19 \\ \bottomrule
    \end{tabular}
    \label{tab:classification_report}
\end{table}

The performance metrics of our proposed AutoML-Drift Compensation (AutoML-DC) model for the proposed drift compensation setting in terms of accuracy, precision, recall, and F1-score are referred to in \Cref{tab:baseline performances}. Each model contributes a certain weight to the learned ensemble. We use meta-learning techniques, automated feature preprocessing techniques, and 
early-stopping and automated ensemble-learning methods. To see the impact of the techniques, we conduct the ablation study in the following subsection.

The simultaneous usage of all AutoML techniques in the AutoML-CD model delivers a 76\% score, which is a performance improvement of 16\% compared to the other benchmarking models.


The AutoML-DC model for the drift compensation setting consists of eight models composed through automated ensemble learning. These eight models comprise five variations of Random Forest models, constituting 85\% of the overall ensemble. Additionally, there are two variants of Multi-Layer Perceptron (MLP) models with early stopping and Tanh activation function, contributing 10\% in total, and one Passive Aggressive algorithm accounting for 50\% of the entire ensemble. This composition indicates that the ensemble is strongly dominated by Random Forests with different regularization strengths, enabling the model to effectively learn and capture both local and global trends in the data. The possibility of AutoML techniques to combine different regularizations of the Random Forest is the most obvious advantage to capture the sensor drift pattern.

If we increase the training dataset size step by step up to 50\% of all initial samples, we observe that a further increase of training data would not be relevant to the model, as it does not increase the score significantly.
If we train the AutoML ensemble with the first 50\% of the data as the training set, the results rise to 96\% F1 score, as seen in~\Cref{tab:classification_report}.

The AutoML optimization pipeline may result in an ensemble of up to 20 models, with our optimization converging to a total of seven models.
This final model, derived through the auto-sklearn optimization, is composed of two MLPs (accounting for 5\% of the decisions each), four random forests (85\% of decisions) and a Passive Aggresive Algorithm (final 5\%).
The Random Forests differ in their minimum number of samples per leaf (i.e. 1, 2, 11 and 19 respectively), with one also using Feature Agglomeration.
Other methods, such as the balancing strategy, classifier selection, data preprocessing techniques, and feature preprocessing methods all contribute to the model's ability to adapt to changing patterns caused by sensor drift.
Moreover, it can be observed that feature preprocessing methods like polynomial transformations and feature agglomeration enhance the model's capability to capture complex patterns like sensor drift. Imputation strategies like mean, median, or most frequent value imputation are also beneficial for handling sensor drift. Tuning additional hyperparameters specific to each algorithm, such as tree depth in Random Forest or the number of nodes per layer in MLP, further enhances the ensemble model's ability to adapt to changing data patterns caused by the inherent sensor drift.

\subsection{Ablation Study}
In this ablation study, we investigate the impact of various components and techniques within our AutoML ensemble framework on the model's performance to compensate for sensor drift. This includes meta-learning techniques, automated feature preprocessing, early stopping, and automated ensemble-learning methods. To thoroughly understand the contribution of each component, we perform this ablation study and remove specific parts of the model to observe changes in performance.

\begin{table}[ht]
    \centering
    \caption{Comparison of Classification Metrics for Different AutoML Ablation Strategies}
    \small
    \setlength{\tabcolsep}{4pt} 
    \begin{tabular}{|c|c|c|c|c|c|c|}
        \hline
        \textbf{Strategy} & \textbf{Class} & \textbf{Prec.} & \textbf{Recall} & \textbf{F1} & \textbf{Acc.} & \textbf{Time/Epoch (s)} \\
        \hline
        \multirow{5}{*}{\shortstack{All \\ Optimiz.}}
        & 1 & 0.65 & 0.94 & 0.77 & \multirow{5}{*}{0.76} & \multirow{5}{*}{116.42} \\
        & 2 & 0.86 & 0.89 & 0.87 & & \\
        & 3 & 0.70 & 0.69 & 0.69 & & \\
        & 4 & 0.57 & 0.38 & 0.46 & & \\
        & 5 & 0.99 & 0.79 & 0.88 & & \\
        \hline
        \multirow{5}{*}{\shortstack{Without \\ Ensemble}}
        & 1 & 0.56 & 0.91 & 0.69 & \multirow{5}{*}{0.68} & \multirow{5}{*}{116.25} \\
        & 2 & 0.79 & 0.75 & 0.77 & & \\
        & 3 & 0.71 & 0.62 & 0.66 & & \\
        & 4 & 0.46 & 0.38 & 0.42 & & \\
        & 5 & 0.99 & 0.65 & 0.78 & & \\
        \hline
        \multirow{5}{*}{\shortstack{Without \\ Preproc.}}
        & 1 & 0.49 & 0.68 & 0.57 & \multirow{5}{*}{0.61} & \multirow{5}{*}{114.07} \\
        & 2 & 0.76 & 0.69 & 0.72 & & \\
        & 3 & 0.45 & 0.57 & 0.50 & & \\
        & 4 & 0.44 & 0.30 & 0.36 & & \\
        & 5 & 0.89 & 0.71 & 0.79 & & \\
        \hline
        \multirow{5}{*}{\shortstack{Without \\ Meta-learn.}}
        & 1 & 0.50 & 0.67 & 0.57 & \multirow{5}{*}{0.63} & \multirow{5}{*}{115.42} \\
        & 2 & 0.76 & 0.71 & 0.73 & & \\
        & 3 & 0.51 & 0.60 & 0.55 & & \\
        & 4 & 0.49 & 0.40 & 0.44 & & \\
        & 5 & 0.91 & 0.72 & 0.81 & & \\
        \hline
    \end{tabular}
    \label{tab:comparison_metrics}
\end{table}

As observable in~\Cref{tab:comparison_metrics}, using all optimization strategies, namely ensemble learning, preprocessing, and meta-learning, in our AutoML-CD yields the best overall performance, with the highest accuracy and balanced precision and recall across most classes while the runtime variations per epoch stay comparable. The removal of any of these strategies results in noticeable drops in performance. Notably, preprocessing and meta-learning are important for improving the classification of more challenging classes like Class 4. These findings suggest that an integrated approach leveraging all available techniques is essential for achieving optimal performance in AutoML.

In all optimization scenarios, the Random Forest algorithm holds the largest share. When training with all possible AutoML optimizations, it constitutes 85\% of the ensemble, compared to 10\% for MLP and 5\% for the Passive Aggressive algorithm. The most complex ensemble in the AutoML-CD model comprises eight models, five of which are Random Forests with varying configurations. In all other settings, only the Random Forest algorithm is used, albeit with different configurations.
Across almost all Random Forest configurations, the Gini criterion is used for measuring split quality, signifying its superiority to other choices. Only when omitting either the preprocessing or the meta-learning steps, criteria other than Gini are used in roughly 25-30\% of all models.


\subsubsection{Impact of Meta-learning}\label{sec:impact_meta_learning}

When omitting the meta-learning module within auto-sklearn, the F1 score drops significantly from 76\% to 63\%.
Compared to the benchmarking models, our performance advantage diminishes to just 3\%. This significant reduction underscores the importance of meta-learning. If we enlengthen the limit of time per run and the time for the task this effect is reduced, but as we compare all models with the same number of epochs per training, this stays an important factor.

Meta-learning involves initializing the hyperparameter optimization algorithm with configurations that have proven effective on previously observed datasets~\cite{AutoML:auto-sklearn}. Its absence would necessitate more computational resources and time to achieve comparable results, making it an important technique for the final AutoML-DC model. 

Meta-learning improves the AutoML-DC efficiency by utilizing prior knowledge from similar tasks to inform hyperparameter configurations. It does so by leveraging meta-features that describe dataset characteristics like dimensionality and class imbalance, allowing for optimized and faster initializations~\cite{AutoML:SMAC}. This leads to accelerated optimization, reduced computational time, and improved model adaptation to sensor drift. In the absence of meta-learning, achieving comparable performance would require significantly more computational resources. Consequently, meta-learning is pivotal for efficient hyperparameter tuning, resulting in a robust and diverse ensemble of models that effectively address the drift compensation task.

\subsubsection{Impact of Ensemble-Learning}

When we omit automated ensemble learning techniques and instead select the best individual model from the ensemble, in this case Random Forest, the F1 score decreases from 76\% to 68\%, which still outperforms all other baselines in~\Cref{tab:baseline performances}.

The found Random Forest utilizes minimal depths of two leaves, with bootstrap always turned off, the criterion remaining as Gini, class balancing being omitted, and the imputation strategy remains at mean. 

\subsubsection{Impact of Preprocessing}

When leaving out the optimization of preprocessing strategies, the F1-score drops around 15\% from 76\% to 61\%. This signifies the highest drop in F1 out of all the ablation studies.

Without feature preprocessing optimization, the number of models with different configurations diminishes again. Here, only two Random Forests remain, one with a minimal sampling leaf depth of 1 and one with a depth of 7 leaves.
Other than that, only the imputation strategy is different, with one using mean and the other using most-frequent. Thus, the number of leaves in the Random Forest and the imputation strategy are the key factors distinguishing the two learned models.

As evidenced by the F1 score, preprocessing is particularly important in the trained AutoML-CD model, as the combination of various preprocessing techniques appears to effectively capture different aspects of drift behavior. Hybrid sensor drift can exhibit a wide range of variances. Preprocessing techniques, such as feature scaling, polynomial feature generation, and feature agglomeration, help normalize and transform the data, making it easier for the models to detect underlying patterns and trends like sensor drift.

\subsection{Comparison of misclassifications}

\begin{figure}
    \centering
    \includegraphics[width=\linewidth]{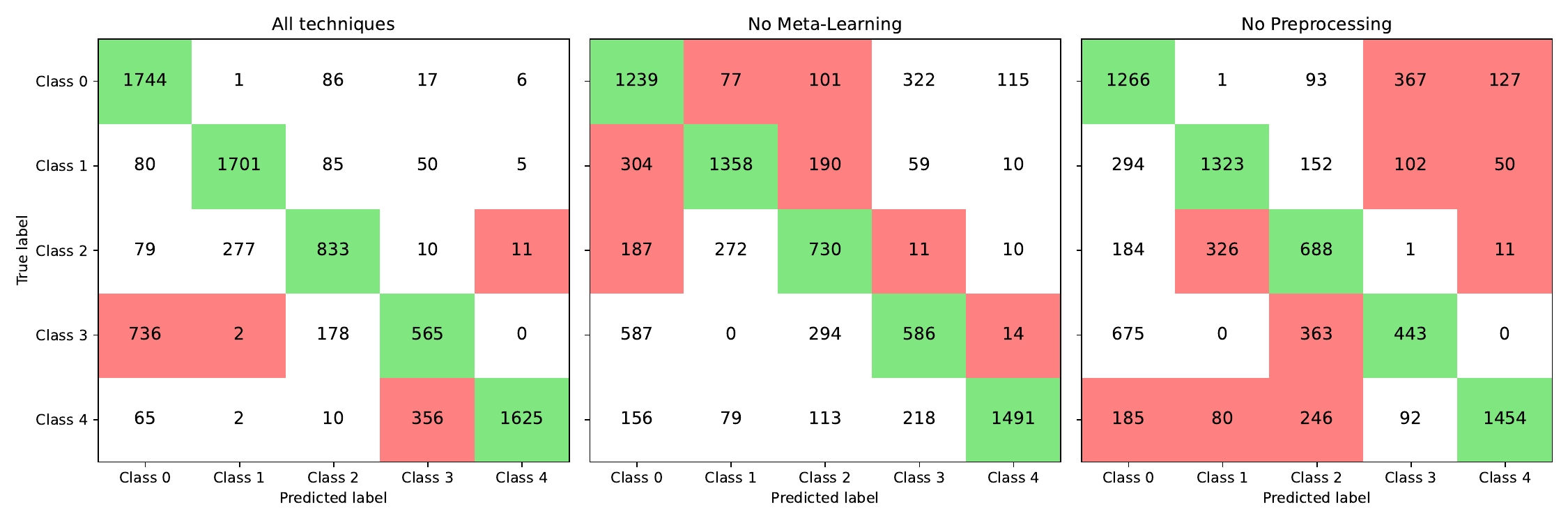}
    \caption{Confusion matrices for all ablation studies, including the pipeline with all techniques, without meta-learning, and without preprocessing.}
    \label{fig:automl_confusion_matrices}
\end{figure}

We may compare the failure cases using~\Cref{fig:automl_confusion_matrices}. Without automatic feature preprocessing, the highest misclassification is made on Class 5 compared to the other strategies. Without meta-learning, the ensemble more frequently incorrectly predicts Class 1.
Class 4 is the most frequently misclassified, pulling down the average of all models, despite the training set being relatively balanced.
This suggests that Class 4 has peculiarities in the data that make it difficult to predict.

The lack of preprocessing might result in the model not capturing the essential patterns in the data. The highest misclassification occurs in Class 5 when preprocessing is not used.
The absence of meta-learning leads to suboptimal model configurations, causing the ensemble to mispredict Class 1 frequently. 
Despite having fewer samples, Class 3 is easier to classify than Class 4. This anomaly suggests that Class 4's data has unique challenges, such as higher variability or less distinct boundaries between classes.
The frequent misclassification of Class 4 indicates underlying complexities within the data. This is stated to be ~\cite{Dennler2021DriftIA} due to a differing sensor drift behaviour of this class in the training data. To keep this information, we chose to leave the multi-class task instead of summoning a normal and an anomalous class like it is frequently used in anomaly detection settings.




\section{Limitations}\label{sec:limitations}

Our study has several limitations that need to be acknowledged. Firstly, the use of a single dataset restricts the generalizability of our findings, as insights derived may be specific to this dataset, but this is caused by the lack of real-world datasets for sensor drift. We avoided using synthetic datasets as they allow for easy re-engineering of known drifts, offering limited real-world applicability despite being solvable with tailored models for Gaussian noise. Furthermore, the predictability of synthetic data does not translate well to real-world scenarios. 
While synthetic datasets with controlled drift patterns are useful for preliminary controlled experiments, they lack the intricate and nuanced variations found in real-world drift phenomena, leading to models achieving higher performance due to their uniform and predictable nature. Consequently, relying on these datasets provides an overly optimistic assessment of model capabilities, rendering them inadequate for thorough validation in practical applications like sensor drift studies, which is why they were excluded from our evaluation.
On the other hand, real-world sensor drift datasets often lack ground truth labels, providing only pre- and post-calibration points, which complicates effective validation and comparison of models. Additionally, data imbalance highly influences our decision-making, which could affect model interpretability. For further understanding of the dataset, the explanation of the dataset imbalance has been moved to the appendix. Future research should focus on the measurement of novel real-time datasets to enhance the robustness and generalizability of the results. Ongoing efforts to obtain more comprehensive datasets could also address the challenge of missing ground truth labels in real-world data. Another limitation of our approach is the challenge of dealing with the complex nature of drift behavior in the data, which complicates explicit drift modeling. If the drift is heavily influenced by unseen factors that hinder predictability, our approach could get worse. This limitation also highlights the conditions under which our method works best and points to situations where additional considerations may be necessary. However, our approach demonstrated notable success with the given dataset, as evidenced by the superior performance of the AutoML-DC model relative to other evaluated methods. AutoML-DC's performance suggests that it can capture significant temporal patterns, despite the presence of baseline drift, short-term, and long-term drift effects. However, to evaluate this, other real-world datasets would be needed.
From the benchmarking results in~\Cref{sec6:experimental_results}, particularly in~\Cref{sec:experiments_anomaly_detection}, it is clear that certain models perform differently based on their ability to adapt to varying drift severities (see the performance discrepancies among models such as Random Forest, SVM, and AutoML-DC).~\Cref{sec:impact_meta_learning} on the impact of meta-learning also highlights how essential prior knowledge is for handling drift effectively. These experimental results collectively suggest that the model's robustness can fluctuate depending on the mode of drift it encounters and should be further investigated.

\section{Conclusion}\label{sec:conclusion}

Adressing sensor drift compensation, we explored the performance of various machine learning models and demonstrated the effectiveness of Automated Machine Learning (AutoML) techniques in drift compensation. The preprocessing steps, including feature scaling and polynomial feature selection, enhanced the ability of the AutoML-DC model to compensate for drift variance. Ensembles of Random Forests with varying regularization strengths learned local and global trends and adapted to significant distributional changes. The ability of AutoML techniques to combine different Random Forest regularizations effectively adapted the model to the significant distributional changes from the initial state to increased sensor drift states. We showed that a model’s capability to compensate for sensor drift also highly depends on the training setting. Thus, we introduced a novel training approach that is more effective than traditional methods. Future research should refine these approaches for unsupervised training to better address real-world sensor drift challenges.
Our anomaly detection-inspired training approach allowed us to identify which models are able to successfully generalize from initial drift-free data to datasets with increased drift. Meanwhile, incremental batch learning validated seamless adaptation, allowing models to evolve with new data streams and maintaining performance amidst continual change.

\section{Acknowledgement}
Thanks to FlowChief GmbH for supporting this study.

\bibliography{references}

\begin{thebibliography}{10}
\expandafter\ifx\csname url\endcsname\relax
  \def\url#1{\texttt{#1}}\fi
\expandafter\ifx\csname urlprefix\endcsname\relax\def\urlprefix{URL }\fi
\expandafter\ifx\csname href\endcsname\relax
  \def\href#1#2{#2} \def\path#1{#1}\fi

\bibitem{margarit2022cross}
J.~M. Margarit-Taul{\'e}, M.~Mart{\'\i}n-Ezquerra, R.~Escud{\'e}-Pujol, C.~Jim{\'e}nez-Jorquera, S.-C. Liu, Cross-compensation of fet sensor drift and matrix effects in the industrial continuous monitoring of ion concentrations, Sensors and Actuators B: Chemical 353 (2022) 131123.

\bibitem{shen2020drift}
J.~Shen, J.~Y. Won, Z.~Chen, Q.~A. Chen, Drift with devil: Security of $\{$Multi-Sensor$\}$ fusion based localization in $\{$High-Level$\}$ autonomous driving under $\{$GPS$\}$ spoofing, in: 29th USENIX security symposium (USENIX Security 20), 2020, pp. 931--948.

\bibitem{wadinger2024adaptable}
M.~Wadinger, M.~Kvasnica, Adaptable and interpretable framework for anomaly detection in scada-based industrial systems, Expert Systems with Applications 246 (2024) 123200.

\bibitem{bourgeois2003use}
W.~Bourgeois, A.-C. Romain, J.~Nicolas, R.~M. Stuetz, The use of sensor arrays for environmental monitoring: interests and limitations, Journal of Environmental Monitoring 5~(6) (2003) 852--860.

\bibitem{holmberg1996drift}
M.~Holmberg, F.~Winquist, I.~Lundstr{\"o}m, F.~Davide, C.~DiNatale, A.~D'Amico, Drift counteraction for an electronic nose, Sensors and Actuators B: Chemical 36~(1-3) (1996) 528--535.

\bibitem{yuyan2015fault}
C.~Yuyan, W.~Jian, X.~Rong, W.~Xinmin, Fault tree analysis of electro-mechanical actuators, in: 2015 34th Chinese Control Conference (CCC), IEEE, 2015, pp. 6392--6396.

\bibitem{godwin2009accuracy}
A.~Godwin, M.~Agnew, J.~Stevenson, Accuracy of inertial motion sensors in static, quasistatic, and complex dynamic motion, Journal of biomechanical engineering 131 (2009) 114501.
\newblock \href {https://doi.org/10.1115/1.4000109} {\path{doi:10.1115/1.4000109}}.

\bibitem{WANG2025115573}
Q.~Wang, J.~Li, S.~Zhang, H.~Tian, S.~Jie, C.~Qu, Z.~Liu, \href{https://www.sciencedirect.com/science/article/pii/S0263224124014581}{A novel temperature drift compensation method based on lstm for nmr sensor}, Measurement 240 (2025) 115573.
\newblock \href {https://doi.org/https://doi.org/10.1016/j.measurement.2024.115573} {\path{doi:https://doi.org/10.1016/j.measurement.2024.115573}}.
\newline\urlprefix\url{https://www.sciencedirect.com/science/article/pii/S0263224124014581}

\bibitem{AHMAD2024115158}
R.~Ahmad, \href{https://www.sciencedirect.com/science/article/pii/S0263224124010431}{Enhanced drift self-calibration of low-cost sensor networks based on cluster and advanced statistical tools}, Measurement 236 (2024) 115158.
\newblock \href {https://doi.org/https://doi.org/10.1016/j.measurement.2024.115158} {\path{doi:https://doi.org/10.1016/j.measurement.2024.115158}}.
\newline\urlprefix\url{https://www.sciencedirect.com/science/article/pii/S0263224124010431}

\bibitem{chandola2009anomaly}
V.~Chandola, A.~Banerjee, V.~Kumar, Anomaly detection: A survey, ACM computing surveys (CSUR) 41~(3) (2009) 1--58.

\bibitem{VERGARA2012320}
A.~Vergara, S.~Vembu, T.~Ayhan, M.~A. Ryan, M.~L. Homer, R.~Huerta, \href{https://www.sciencedirect.com/science/article/pii/S0925400512002018}{Chemical gas sensor drift compensation using classifier ensembles}, Sensors and Actuators B: Chemical 166-167 (2012) 320--329.
\newblock \href {https://doi.org/https://doi.org/10.1016/j.snb.2012.01.074} {\path{doi:https://doi.org/10.1016/j.snb.2012.01.074}}.
\newline\urlprefix\url{https://www.sciencedirect.com/science/article/pii/S0925400512002018}

\bibitem{Definition_Drift_Adaptation}
F.~Bayram, B.~S. Ahmed, A.~Kassler, \href{https://doi.org/10.1016/j.knosys.2022.108632}{From concept drift to model degradation: An overview on performance-aware drift detectors}, Knowl. Based Syst. 245 (2022) 108632.
\newblock \href {https://doi.org/10.1016/J.KNOSYS.2022.108632} {\path{doi:10.1016/J.KNOSYS.2022.108632}}.
\newline\urlprefix\url{https://doi.org/10.1016/j.knosys.2022.108632}

\bibitem{SUAREZCETRULO2023118934}
A.~L. Suárez-Cetrulo, D.~Quintana, A.~Cervantes, \href{https://www.sciencedirect.com/science/article/pii/S0957417422019522}{A survey on machine learning for recurring concept drifting data streams}, Expert Systems with Applications 213 (2023) 118934.
\newblock \href {https://doi.org/https://doi.org/10.1016/j.eswa.2022.118934} {\path{doi:https://doi.org/10.1016/j.eswa.2022.118934}}.
\newline\urlprefix\url{https://www.sciencedirect.com/science/article/pii/S0957417422019522}

\bibitem{TangWangWuChenPangSunFengWang+2023}
Y.~Tang, Y.~Wang, D.~Wu, M.~Chen, L.~Pang, J.~Sun, W.~Feng, X.~Wang, \href{https://doi.org/10.1515/rams-2023-0347}{Exploring temperature-resilient recycled aggregate concrete with waste rubber: An experimental and multi-objective optimization analysis}, REVIEWS ON ADVANCED MATERIALS SCIENCE 62~(1) (2023) 20230347 [cited 2025-01-09].
\newblock \href {https://doi.org/doi:10.1515/rams-2023-0347} {\path{doi:doi:10.1515/rams-2023-0347}}.
\newline\urlprefix\url{https://doi.org/10.1515/rams-2023-0347}

\bibitem{TangWangWuLiuZhangZhuChenSunWang+2022+849+872}
Y.~Tang, Y.~Wang, D.~Wu, Z.~Liu, H.~Zhang, M.~Zhu, Z.~Chen, J.~Sun, X.~Wang, \href{https://doi.org/10.1515/rams-2022-0274}{An experimental investigation and machine learning-based prediction for seismic performance of steel tubular column filled with recycled aggregate concrete}, REVIEWS ON ADVANCED MATERIALS SCIENCE 61~(1) (2022) 849--872 [cited 2025-01-09].
\newblock \href {https://doi.org/doi:10.1515/rams-2022-0274} {\path{doi:doi:10.1515/rams-2022-0274}}.
\newline\urlprefix\url{https://doi.org/10.1515/rams-2022-0274}

\bibitem{Schaller2023LiquorHGNNAH}
M.~Schaller, M.~Steininger, A.~Dulny, D.~Schl{\"o}r, A.~Hotho, \href{https://api.semanticscholar.org/CorpusID:262103439}{Liquor-hgnn: A heterogeneous graph neural network for leakage detection in water distribution networks}, in: Lernen, Wissen, Daten, Analysen, 2023, pp. 454--469.
\newline\urlprefix\url{https://api.semanticscholar.org/CorpusID:262103439}

\bibitem{schaller2024modeconvnovelconvolutiondistinguishing}
M.~Schaller, D.~Schlör, A.~Hotho, \href{https://arxiv.org/abs/2407.00140}{Modeconv: A novel convolution for distinguishing anomalous and normal structural behavior} (2024).
\newblock \href {http://arxiv.org/abs/2407.00140} {\path{arXiv:2407.00140}}.
\newline\urlprefix\url{https://arxiv.org/abs/2407.00140}

\bibitem{AutoML:SMAC}
M.~Lindauer, K.~Eggensperger, M.~Feurer, A.~Biedenkapp, D.~Deng, C.~Benjamins, T.~Ruhkopf, R.~Sass, F.~Hutter, \href{http://jmlr.org/papers/v23/21-0888.html}{{SMAC3:} {A} versatile bayesian optimization package for hyperparameter optimization}, J. Mach. Learn. Res. 23 (2022) 54:1--54:9.
\newline\urlprefix\url{http://jmlr.org/papers/v23/21-0888.html}

\bibitem{AutoML:auto-sklearn}
M.~Feurer, A.~Klein, K.~Eggensperger, J.~T. Springenberg, M.~Blum, F.~Hutter, Efficient and robust automated machine learning, in: Advances in Neural Information Processing Systems 28: Annual Conference on Neural Information Processing Systems 2015, December 7-12, 2015, Montreal, Quebec, Canada, 2015, pp. 2962--2970.

\bibitem{AutoML:CASH}
C.~Thornton, F.~Hutter, H.~H. Hoos, K.~Leyton{-}Brown, \href{https://doi.org/10.1145/2487575.2487629}{Auto-weka: combined selection and hyperparameter optimization of classification algorithms}, in: I.~S. Dhillon, Y.~Koren, R.~Ghani, T.~E. Senator, P.~Bradley, R.~Parekh, J.~He, R.~L. Grossman, R.~Uthurusamy (Eds.), The 19th {ACM} {SIGKDD} International Conference on Knowledge Discovery and Data Mining, {KDD} 2013, Chicago, IL, USA, August 11-14, 2013, {ACM}, 2013, pp. 847--855.
\newblock \href {https://doi.org/10.1145/2487575.2487629} {\path{doi:10.1145/2487575.2487629}}.
\newline\urlprefix\url{https://doi.org/10.1145/2487575.2487629}

\bibitem{AutoML:NAS}
B.~Zoph, Q.~V. Le, \href{https://arxiv.org/abs/1611.01578}{Neural architecture search with reinforcement learning} (2017).
\newblock \href {http://arxiv.org/abs/1611.01578} {\path{arXiv:1611.01578}}.
\newline\urlprefix\url{https://arxiv.org/abs/1611.01578}

\bibitem{artursson2000drift}
T.~Artursson, T.~Ekl{\"o}v, I.~Lundstr{\"o}m, P.~M{\aa}rtensson, M.~Sj{\"o}str{\"o}m, M.~Holmberg, Drift correction for gas sensors using multivariate methods, Journal of chemometrics 14~(5-6) (2000) 711--723.

\bibitem{DINATALE2002158}
C.~{Di Natale}, E.~Martinelli, A.~D’Amico, \href{https://www.sciencedirect.com/science/article/pii/S0925400501010012}{Counteraction of environmental disturbances of electronic nose data by independent component analysis}, Sensors and Actuators B: Chemical 82~(2) (2002) 158--165.
\newblock \href {https://doi.org/https://doi.org/10.1016/S0925-4005(01)01001-2} {\path{doi:https://doi.org/10.1016/S0925-4005(01)01001-2}}.
\newline\urlprefix\url{https://www.sciencedirect.com/science/article/pii/S0925400501010012}

\bibitem{DICARLO20111594}
S.~{Di Carlo}, M.~Falasconi, E.~Sanchez, A.~Scionti, G.~Squillero, A.~Tonda, \href{https://www.sciencedirect.com/science/article/pii/S0167865511001760}{Increasing pattern recognition accuracy for chemical sensing by evolutionary based drift compensation}, Pattern Recognition Letters 32~(13) (2011) 1594--1603.
\newblock \href {https://doi.org/https://doi.org/10.1016/j.patrec.2011.05.019} {\path{doi:https://doi.org/10.1016/j.patrec.2011.05.019}}.
\newline\urlprefix\url{https://www.sciencedirect.com/science/article/pii/S0167865511001760}

\bibitem{HUI2003354}
D.~hui, L.~Jun-hua, S.~Zhong-ru, \href{https://www.sciencedirect.com/science/article/pii/S0925400503005690}{Drift reduction of gas sensor by wavelet and principal component analysis}, Sensors and Actuators B: Chemical 96~(1) (2003) 354--363.
\newblock \href {https://doi.org/https://doi.org/10.1016/S0925-4005(03)00569-0} {\path{doi:https://doi.org/10.1016/S0925-4005(03)00569-0}}.
\newline\urlprefix\url{https://www.sciencedirect.com/science/article/pii/S0925400503005690}

\bibitem{PADILLA201028}
M.~Padilla, A.~Perera, I.~Montoliu, A.~Chaudry, K.~Persaud, S.~Marco, \href{https://www.sciencedirect.com/science/article/pii/S0169743909001877}{Drift compensation of gas sensor array data by orthogonal signal correction}, Chemometrics and Intelligent Laboratory Systems 100~(1) (2010) 28--35.
\newblock \href {https://doi.org/https://doi.org/10.1016/j.chemolab.2009.10.002} {\path{doi:https://doi.org/10.1016/j.chemolab.2009.10.002}}.
\newline\urlprefix\url{https://www.sciencedirect.com/science/article/pii/S0169743909001877}

\bibitem{kohonen}
T.~Kohonen, The self-organizing map, Proceedings of the IEEE 78~(9) (1990) 1464--1480.
\newblock \href {https://doi.org/10.1109/5.58325} {\path{doi:10.1109/5.58325}}.

\bibitem{Falasconi12}
S.~D. Carlo, M.~Falasconi, \href{https://doi.org/10.5772/33411}{Drift correction methods for gas chemical sensors in artificial olfaction systems: Techniques and challenges}, in: W.~Wang (Ed.), Advances in Chemical Sensors, IntechOpen, Rijeka, 2012, Ch.~14, pp. 304--326.
\newblock \href {https://doi.org/10.5772/33411} {\path{doi:10.5772/33411}}.
\newline\urlprefix\url{https://doi.org/10.5772/33411}

\bibitem{HOLMBERG1997185}
M.~Holmberg, F.~A. Davide, C.~{Di Natale}, A.~D'Amico, F.~Winquist, I.~Lundström, \href{https://www.sciencedirect.com/science/article/pii/S0925400597803358}{Drift counteraction in odour recognition applications: lifelong calibration method}, Sensors and Actuators B: Chemical 42~(3) (1997) 185--194.
\newblock \href {https://doi.org/https://doi.org/10.1016/S0925-4005(97)80335-8} {\path{doi:https://doi.org/10.1016/S0925-4005(97)80335-8}}.
\newline\urlprefix\url{https://www.sciencedirect.com/science/article/pii/S0925400597803358}

\bibitem{AutoML:ensemble_selection}
R.~Caruana, A.~Niculescu{-}Mizil, G.~Crew, A.~Ksikes, \href{https://doi.org/10.1145/1015330.1015432}{Ensemble selection from libraries of models}, in: C.~E. Brodley (Ed.), Machine Learning, Proceedings of the Twenty-first International Conference {(ICML} 2004), Banff, Alberta, Canada, July 4-8, 2004, Vol.~69 of {ACM} International Conference Proceeding Series, {ACM}, 2004, p.~18.
\newblock \href {https://doi.org/10.1145/1015330.1015432} {\path{doi:10.1145/1015330.1015432}}.
\newline\urlprefix\url{https://doi.org/10.1145/1015330.1015432}

\bibitem{de2016benchmark}
B.~de~Bruijn, T.~A. Nguyen, D.~Bucur, K.~Tei, \href{https://doi.org/10.5220/0005637901850195}{Benchmark datasets for fault detection and classification in sensor data}, in: A.~Ahrens, O.~Postolache, C.~Benavente{-}Peces (Eds.), {SENSORNETS} 2016 - Proceedings of the 5th International Confererence on Sensor Networks, Rome, Italy, February 19-21, 2016, SciTePress, 2016, pp. 185--195.
\newblock \href {https://doi.org/10.5220/0005637901850195} {\path{doi:10.5220/0005637901850195}}.
\newline\urlprefix\url{https://doi.org/10.5220/0005637901850195}

\bibitem{Dennler2021DriftIA}
N.~Dennler, S.~Rastogi, J.~Fonollosa, A.~van Schaik, M.~Schmuker, \href{https://api.semanticscholar.org/CorpusID:237213501}{Drift in a popular metal oxide sensor dataset reveals limitations for gas classification benchmarks}, Sensors and Actuators B: Chemical (2021).
\newline\urlprefix\url{https://api.semanticscholar.org/CorpusID:237213501}

\bibitem{Lu_2018}
J.~Lu, A.~Liu, F.~Dong, F.~Gu, J.~Gama, G.~Zhang, \href{http://dx.doi.org/10.1109/TKDE.2018.2876857}{Learning under concept drift: A review}, IEEE Transactions on Knowledge and Data Engineering (2018) 1–1\href {https://doi.org/10.1109/tkde.2018.2876857} {\path{doi:10.1109/tkde.2018.2876857}}.
\newline\urlprefix\url{http://dx.doi.org/10.1109/TKDE.2018.2876857}

\bibitem{powers2020evaluationprecisionrecallfmeasure}
D.~M.~W. Powers, \href{https://arxiv.org/abs/2010.16061}{Evaluation: from precision, recall and f-measure to roc, informedness, markedness and correlation} (2020).
\newblock \href {http://arxiv.org/abs/2010.16061} {\path{arXiv:2010.16061}}.
\newline\urlprefix\url{https://arxiv.org/abs/2010.16061}

\bibitem{hastie}
T.~Hastie, R.~Tibshirani, J.~Friedman, J.~Franklin, The elements of statistical learning: Data mining, inference, and prediction, Math. Intell. 27 (2004) 83--85.
\newblock \href {https://doi.org/10.1007/BF02985802} {\path{doi:10.1007/BF02985802}}.

\bibitem{Fawcett}
T.~Fawcett, Introduction to roc analysis, Pattern Recognition Letters 27 (2006) 861--874.
\newblock \href {https://doi.org/10.1016/j.patrec.2005.10.010} {\path{doi:10.1016/j.patrec.2005.10.010}}.

\bibitem{ijaz2020recursive}
M.~Ijaz, A.~ur~Rehman, M.~Hamdi, A.~Bermak, Recursive feature elimination with random forest classifier for compensation of small scale drift in gas sensors, in: 2020 IEEE International Symposium on Circuits and Systems (ISCAS), IEEE, 2020, pp. 1--5.

\bibitem{habib2019classification}
M.~M. Habib, A.~Rodan, A.~Alazzam, A classification model for gas drift problem, in: 2019 Sixth HCT Information Technology Trends (ITT), IEEE, 2019, pp. 172--176.

\bibitem{pareek2021smart}
V.~Pareek, R.~Prajesh, S.~Chaudhury, S.~Singh, Smart gas sensing using single mos gas sensor with adaptive gradient boosting, in: 2021 Joint 10th International Conference on Informatics, Electronics \& Vision (ICIEV) and 2021 5th International Conference on Imaging, Vision \& Pattern Recognition (icIVPR), IEEE, 2021, pp. 1--7.

\bibitem{zhao2019sensor}
X.~Zhao, P.~Li, K.~Xiao, X.~Meng, L.~Han, C.~Yu, Sensor drift compensation based on the improved lstm and svm multi-class ensemble learning models, Sensors 19~(18) (2019) 3844.

\bibitem{saeed2020machine}
U.~Saeed, S.~U. Jan, Y.-D. Lee, I.~Koo, Machine learning-based real-time sensor drift fault detection using raspberry pi, in: 2020 International Conference on Electronics, Information, and Communication (ICEIC), IEEE, 2020, pp. 1--7.

\bibitem{chaudhuri2020attention}
T.~Chaudhuri, M.~Wu, Y.~Zhang, P.~Liu, X.~Li, An attention-based deep sequential gru model for sensor drift compensation, IEEE Sensors Journal 21~(6) (2020) 7908--7917.

\bibitem{jana2022cnn}
D.~Jana, J.~Patil, S.~Herkal, S.~Nagarajaiah, L.~Duenas-Osorio, Cnn and convolutional autoencoder (cae) based real-time sensor fault detection, localization, and correction, Mechanical Systems and Signal Processing 169 (2022) 108723.

\bibitem{zamansky4768947swcnt}
K.~K. Zamansky, F.~Fedorov, S.~Shandakov, M.~Chetyrkina, A.~G. Nasibulin, A swcnt-based free-standing gas sensor for selective recognition of toxic and flammable gases under thermal cycling protocols, Available at SSRN 4768947 (2024).

\bibitem{lin2019concept}
C.-C. Lin, D.-J. Deng, C.-H. Kuo, L.~Chen, Concept drift detection and adaption in big imbalance industrial iot data using an ensemble learning method of offline classifiers, IEEE Access 7 (2019) 56198--56207.

\bibitem{fan2023fault}
C.~Fan, C.~Li, Y.~Peng, Y.~Shen, G.~Cao, S.~Li, Fault diagnosis of vibration sensors based on triage loss function-improved xgboost, Electronics 12~(21) (2023) 4442.

\bibitem{sarnovsky2021adaptive}
M.~Sarnovsky, J.~Marcinko, Adaptive bagging methods for classification of data streams with concept drift, Acta Polytechnica Hungarica 18~(3) (2021) 47--63.

\bibitem{adhikari2014multiple}
S.~Adhikari, S.~Saha, Multiple classifier combination technique for sensor drift compensation using ann \& knn, in: 2014 IEEE International Advance Computing Conference (IACC), IEEE, 2014, pp. 1184--1189.

\bibitem{9756340}
H.~Chen, D.~Huo, J.~Zhang, Gas recognition in e-nose system: A review, IEEE Transactions on Biomedical Circuits and Systems 16~(2) (2022) 169--184.
\newblock \href {https://doi.org/10.1109/TBCAS.2022.3166530} {\path{doi:10.1109/TBCAS.2022.3166530}}.

\bibitem{ngo2019fence}
P.~C. Ngo, A.~A. Winarto, C.~K.~L. Kou, S.~Park, F.~Akram, H.~K. Lee, Fence gan: Towards better anomaly detection, in: 2019 IEEE 31St International Conference on tools with artificial intelligence (ICTAI), IEEE, 2019, pp. 141--148.

\bibitem{lian2023anomaly}
Y.~Lian, Y.~Geng, T.~Tian, Anomaly detection method for multivariate time series data of oil and gas stations based on digital twin and mtad-gan, Applied Sciences 13~(3) (2023) 1891.

\bibitem{raturi2023novel}
R.~Raturi, A.~Kumar, N.~Vyas, V.~Dutt, A novel approach for anomaly detection in time-series data using generative adversarial networks, in: 2023 International Conference on Sustainable Computing and Smart Systems (ICSCSS), IEEE, 2023, pp. 1352--1357.

\bibitem{noor2023generative}
S.~Noor, S.~U. Bazai, M.~I. Ghafoor, S.~Marjan, S.~Akram, F.~Ali, Generative adversarial networks for anomaly detection: A systematic literature review, in: 2023 4th International Conference on Computing, Mathematics and Engineering Technologies (iCoMET), IEEE, 2023, pp. 1--6.

\bibitem{goutte2005probabilistic}
C.~Goutte, E.~Gaussier, A probabilistic interpretation of precision, recall and f-score, with implication for evaluation, in: European conference on information retrieval, Springer, 2005, pp. 345--359.

\bibitem{li2018decision}
Y.~Li, L.~Ding, X.~Gao, On the decision boundary of deep neural networks, arXiv preprint arXiv:1808.05385 (2018).

\end{thebibliography}
\newpage
\appendix

\setcounter{table}{0}
\renewcommand{\thetable}{A\arabic{table}}
\setcounter{figure}{0}
\renewcommand{\thefigure}{A\arabic{figure}}

\section{Dataset Details}
The dataset utilized in this study was curated by Vergara et al.~\cite{VERGARA2012320}. The sensor drift dataset comprises a collection of sensor responses obtained using a sixteen-screen-printed array of commercially available metal-oxide semiconductor gas sensors. The sensors were incorporated into a 60 ml-volume test chamber.

The experimental setup involved injecting various odorants of interest, such as ammonia, acetaldehyde, acetone, ethylene, ethanol, and toluene, into the test chamber in gaseous form. A computer-controlled continuous flow system was employed to regulate the conveyance of chemical compounds at desired concentrations to the sensing chamber. The system featured three digital mass flow controllers (MFCs), each with different maximum flow levels (200, 100, and 20 ml/min, $\pm$ 1\% accuracy). These MFCs were connected to pressurized gas cylinders containing either the carrier gas or the chemical analytes to be measured, diluted in dry air.

To maintain a consistent moisture level of 10\% (measured at 25 $\pm$ 1 °C) throughout the measurements, synthetic dry air was employed as the background gas for all measurements. The total flow rate across the sensing chamber was set to 200 ml/min and kept constant for the entire measurement process. The gas sensor array's response was recorded at an operating temperature of 400 °C, achieved through a built-in heater driven by an external DC voltage source set at 5 V. Preceding the experimental procedures, the sensors underwent a pre-heating phase for several days to ensure reproducible response patterns.

The acquired sensor responses, in the form of the resistance across the active layer of each sensor, constitute a 16-channel time series sequence for each measurement with a 100 Hz sampling rate. Each measurement cycle took at least 300 seconds. The data acquisition board collected the sensor data and controlled the analog voltage signal to each sensor heater. 

The dataset comprises 13,910 recordings collected over 36 months, with each gas type-concentration pair sampled. There was also an intentional inclusion of a 5-month gap where the sensors were powered off, causing contamination~\cite{VERGARA2012320}. 

\subsection{Baseline drift}\label{sec:appendix_a1}
\begin{figure*}[ht]
    \centering
    \includegraphics[width=0.99\textwidth]{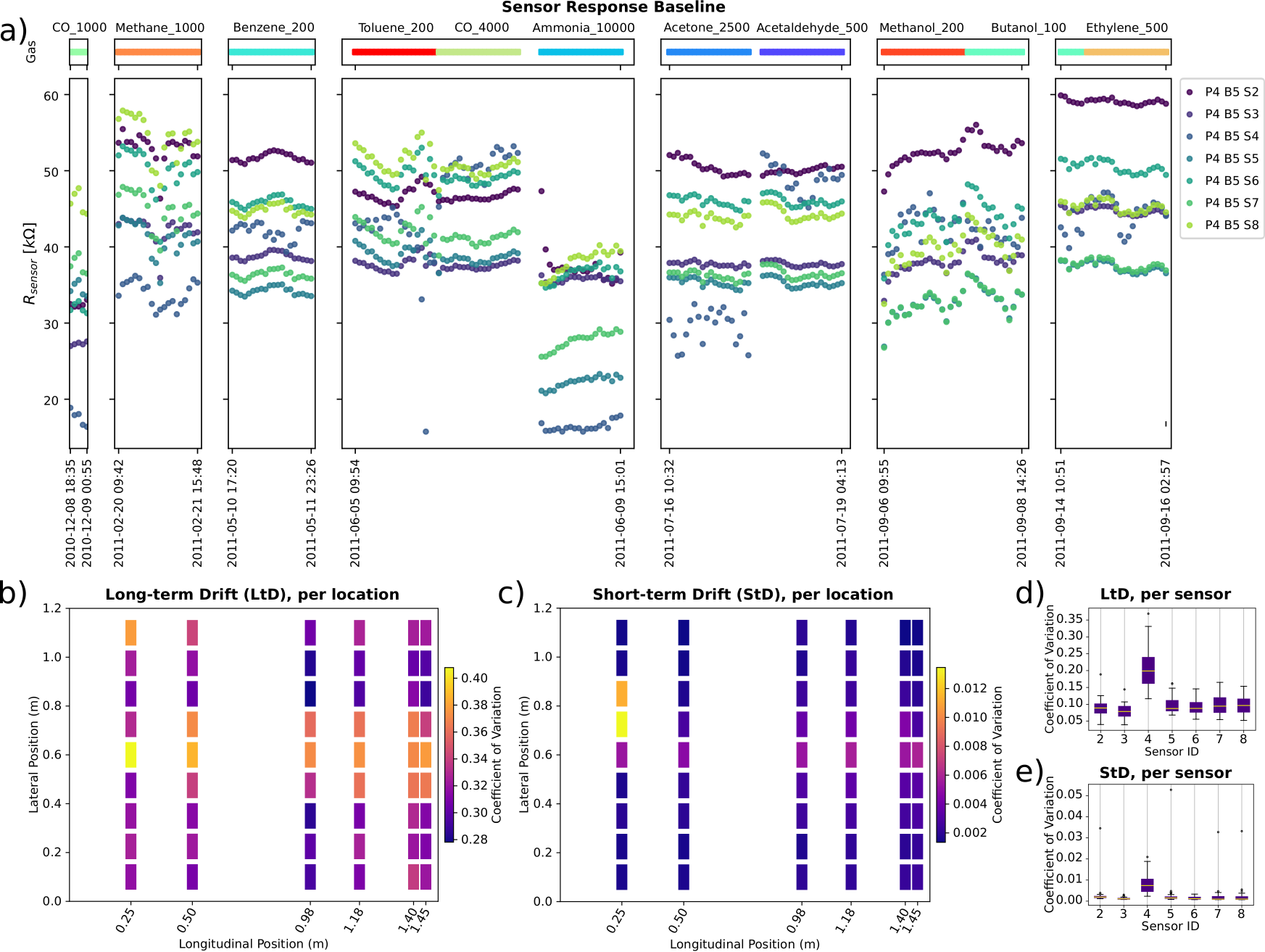}
    \caption{a) Baseline for each sensor and trial, with dots indicating mean sensor resistance before gas release (20 s). The top row shows gas identity and concentration (ppm). b)-e) Local baseline variation analysis using coefficient of variation for spatial wind tunnel locations (b, c) and sensor boards (d, e). b) and d) display long-term baseline variation over the 16-month experiment, while c) and e) show averaged within-trial, short-term baseline variation. Data was obtained with 0.21 m s-1 wind flow speed and 6 V hotplate voltage, considering all ten gases and sensors 2–8. For a), only location 4 and board 5 were considered \cite{Dennler2021DriftIA}.}
    \label{fig:dennler_2}
\end{figure*}

Dennler et al. \cite{Dennler2021DriftIA} also examined the sensor baseline, defined as readings before gas release. The following \Cref{fig:dennler_2} displays the trial-wise average baseline values for a fixed sensor board location, operating conditions, and airflow velocity. The dots represent mean sensor resistance before gas release (20 s), showing significant variations over time. Long-term drift is evident as discontinuities between recording sessions, often correlating with gas identity due to batched gas presentations. Additionally, substantial baseline drift occurs within some recording sessions.

\subsection{Batch distributions and dataset structure}\label{sec:appendix_a2}

It can be stated that the distributions vary widely due to sensor drift effects over time. 
\Cref{gas_overview} provides the gas names corresponding to the six distinct classes Ammonia, Acetaldehyde, Acetone, Ethylene, Ethanol and Toluene, along with the interval of concentration levels for each gas in ppmv as well as the number of samples per class. 

\begin{table*}
\caption{Overview of the six classes of gases, their concentration levels (minimum and maximum), and numbers of samples per class.}
\centering
\begin{tabular}{|c|c|c|c|c|c|c|}
\hline
Gas & Ammon. & Acetal. & Acet. & Ethyl. & Etha. & Tolu. \\
\hline
concent. & & & & & &  \\
(ppmv)& \multirow{-2}{*}{50,1000} & \multirow{-2}{*}{5,500} & \multirow{-2}{*}{12,1000} & \multirow{-2}{*}{10,300} & \multirow{-2}{*}{10,600} & \multirow{-2}{*}{10,100} \\
\hline
samples & 2565 & 2926 & 1641 & 1936 & 3009 & 1833 \\
\hline
\end{tabular}
\label{gas_overview}
\end{table*}

The data has been structured into ten batches for processing purposes, each containing the number of measurements per class and month as indicated in \Cref{tab:monthly_examples}.

For the exact distribution of the gases among the samples and batches 
\Cref{tab:monthly_examples} is inserted. Classes one to six are in the same order as in \Cref{gas_overview}. 

\begin{table*}
\centering
\caption{Month IDs and number of samples per class (1-6 denoting the gases) and B-ID symbolising Batches in tabular overview.}
\begin{tabular}{r|c|c|c|c|c|c|c|c|}
\cline{2-9}
B-ID & Month ID & 1 & 2 & 3 & 4 & 5 & 6 & Total \\
\cline{2-9}
\multirow{2}{*}{1}\ldelim\{{2}{1.5mm} & month1 & 76 & 0 & 0 & 88 & 84 & 0 & 248 \\
& month2 & 7 & 30 & 70 & 10 & 6 & 74 & 197 \\
\cdashline{2-9}
\multirow{5}{*}{2}\ldelim\{{5}{1.5mm} & month3 & 0 & 0 & 7 & 140 & 70 & 0 & 217 \\
&month4 & 0 & 4 & 0 & 170 & 82 & 5 & 261 \\
&month8 & 0 & 0 & 0 & 20 & 0 & 0 & 20 \\
&month9 & 0 & 0 & 0 & 4 & 11 & 0 & 15 \\
&month10 & 100 & 105 & 525 & 0 & 1 & 0 & 731 \\
\cdashline{2-9}
\multirow{3}{*}{3}\ldelim\{{3}{1.5mm}&month11 & 0 & 0 & 0 & 146 & 360 & 0 & 506 \\
&month12 & 0 & 192 & 0 & 334 & 0 & 0 & 526 \\
&month13 & 216 & 48 & 275 & 10 & 5 & 0 & 554 \\
\cdashline{2-9}
\multirow{2}{*}{4}\ldelim\{{2}{1.5mm}&month14 & 0 & 18 & 0 & 43 & 52 & 0 & 113 \\
&month15 & 12 & 12 & 12 & 0 & 12 & 0 & 48 \\
\cdashline{2-9}
\multirow{1}{*}{5}\ldelim\{{1}{1.5mm}&month16 & 20 & 46 & 63 & 40 & 28 & 0 & 197 \\
\cdashline{2-9}
\multirow{4}{*}{6}\ldelim\{{4}{1.5mm}&month17 & 0 & 0 & 0 & 20 & 0 & 0 & 20 \\
&month18 & 0 & 0 & 0 & 3 & 0 & 0 & 3 \\
&month19 & 110 & 29 & 140 & 100 & 264 & 9 & 652 \\
&month20 & 0 & 0 & 466 & 451 & 250 & 458 & 1625 \\
\cdashline{2-9}
\multirow{1}{*}{7}\ldelim\{{1}{1.5mm}&month21 & 360 & 744 & 630 & 662 & 649 & 568 & 3613 \\
\cdashline{2-9}
\multirow{2}{*}{8}\ldelim\{{2}{1.5mm}&month22 & 25 & 15 & 123 & 0 & 0 & 0 & 163 \\
&month23 & 15 & 18 & 20 & 30 & 30 & 18 & 131 \\
\cdashline{2-9}
\multirow{2}{*}{9}\ldelim\{{2}{1.5mm}&month24 & 0 & 25 & 28 & 0 & 0 & 1 & 54 \\
&month30 & 100 & 50 & 50 & 55 & 61 & 100 & 416 \\
\cdashline{2-9}
\multirow{1}{*}{10}\ldelim\{{1}{1.5mm}&month36 & 600 & 600 & 600 & 600 & 600 & 600 & 3600 \\
\cline{2-9}
\end{tabular}
\label{tab:monthly_examples}
\end{table*}

\end{document}